\documentclass[11pt]{article}
\PassOptionsToPackage{table}{xcolor}
\usepackage{acl}
\usepackage{times,latexsym}
\usepackage[T1]{fontenc}
\usepackage[utf8]{inputenc}
\usepackage{microtype}
\usepackage{graphicx,amsmath,amssymb}
\usepackage{booktabs,multirow,siunitx,array,makecell}
\usepackage{placeins,fvextra,seqsplit,adjustbox,tabularx}
\usepackage{xcolor}
\usepackage{enumitem}
\definecolor{tactHeat}{RGB}{43,108,176}

\definecolor{tactProtocolFill}{RGB}{244,246,248}
\definecolor{tactViewSep}{RGB}{204,214,224}
\definecolor{tactRowFill}{RGB}{224,238,249}
\definecolor{tactGainFill}{RGB}{248,240,226}
\definecolor{tactStdText}{RGB}{90,108,125}
\newcommand{\tactstd}[1]{{\fontsize{6.5}{7}\selectfont\textcolor{tactStdText}{$\pm$\,#1}}}

\definecolor{protocolbg}{RGB}{246,249,253}
\definecolor{tactrow}{RGB}{234,243,255}
\definecolor{gainrowfill}{RGB}{234,243,255}
\definecolor{gaintext}{RGB}{28,79,150}
\definecolor{viewsep}{RGB}{205,211,218}

\newcolumntype{C}[1]{>{\centering\arraybackslash}m{#1}}
\newcolumntype{L}[1]{>{\raggedright\arraybackslash}m{#1}}

\newcommand{\methodname}{\textsc{TACT}}

\title{From Relevance to Validity: Zero-Shot Tool Selection
via Functional and Contextual Validation}

\author{
\normalfont\mdseries
Xiaoxuan Dong\textsuperscript{1} \quad
Jin Zhang\textsuperscript{2} \quad
Zichen Yuan\textsuperscript{3} \quad
Yue Zhou\textsuperscript{2} \quad
Zihao Rao\textsuperscript{3}
\\
\normalfont\mdseries
Dexu Yu\textsuperscript{4} \quad
Tingyong Wu\textsuperscript{1} \quad
Wenhao Deng\textsuperscript{5} \quad
Junchen Fu\textsuperscript{5} \quad
Pengwei Sun\textsuperscript{6}
\\
\normalfont\mdseries
Yang Ran\textsuperscript{7} \quad
Bo Lan\textsuperscript{7} \quad
Yongxin Ni\textsuperscript{8} \quad
Youhua Li\textsuperscript{3,*}
\\[0.7em]
\small\normalfont\mdseries
\textsuperscript{1}UESTC \quad
\textsuperscript{2}LZU \quad
\textsuperscript{3}CityUHK \quad
\textsuperscript{4}Northwestern \quad
\textsuperscript{5}UofG \quad
\textsuperscript{6}Stanford \quad
\textsuperscript{7}Fenz AI \quad
\textsuperscript{8}NUS
\\[0.4em]
\small\normalfont\mdseries
\textsuperscript{*}\texttt{Corresponding: youhuali2-c@my.cityu.edu.hk}
}

\begin{document}
\maketitle

\begin{abstract}
Large language model (LLM) agents increasingly face tasks from unseen domains and execution contexts, making zero-shot tool selection essential for identifying tools that can satisfy task requirements.
Existing dominant methods mainly rely on relevance ranking, but semantic matching alone cannot ensure that a tool is functionally compatible or remains applicable in a new context.
We preliminarily identify a \emph{transfer-validity gap}, where semantically relevant tools may still fail to satisfy functional or contextual requirements after transfer.
To address this gap, we propose \methodname{}, a progressive validation framework that complements semantic relevance with endpoint-grounded functional compatibility and task-conditioned contextual evidence.
Across cross-domain and cross-dataset transfer, \methodname{} consistently improves capability ranking, with larger gains under functional or contextual ambiguity.
Ablation results further confirm the complementary contributions of functional and contextual validation.
\end{abstract}

\section{Introduction}
\label{sec:introduction}

As large language models increasingly interact with external tools, APIs,
and services, agents must select suitable capabilities from growing libraries
\citep{yao2023react,schick2023toolformer}.
This problem becomes harder under distribution shift, where tasks may come
from unseen domains or execution settings and target-side supervision is
unavailable. Reliable zero-shot selection therefore requires transferring
previously learned task--capability relations to new contexts.

\begin{figure}[!t]
\centering
\includegraphics[width=\columnwidth]{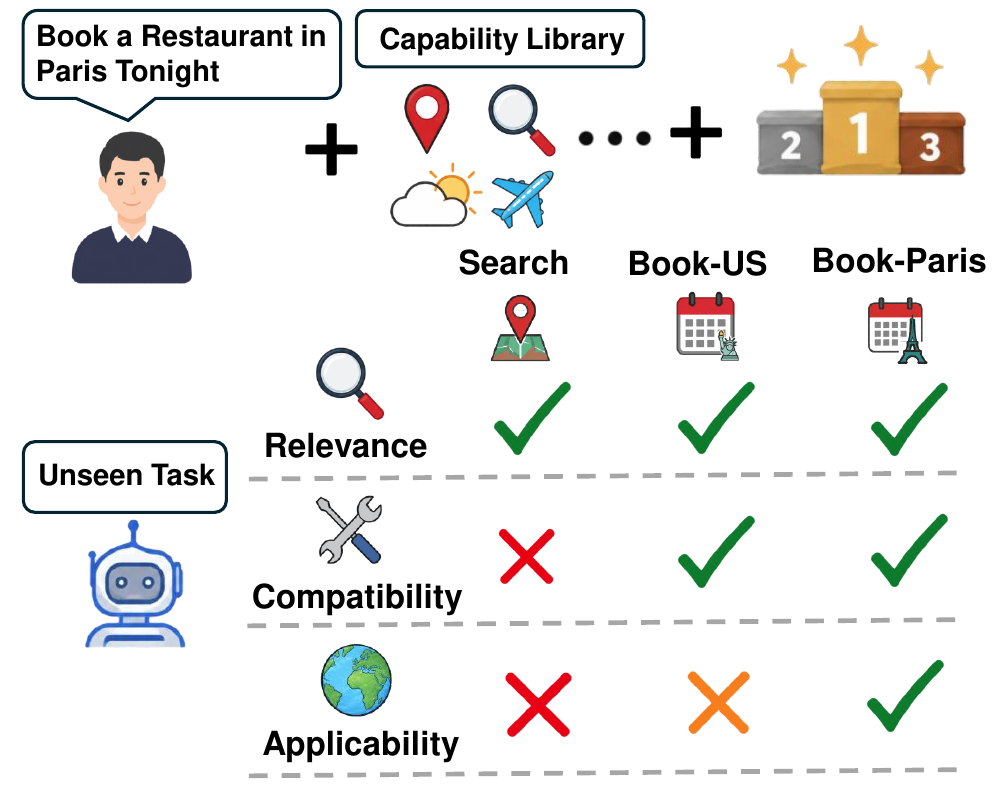}
\caption{
Zero-shot capability transfer involves three questions:
whether a candidate is relevant, whether its endpoints can
satisfy the task, and whether source-side support remains
applicable to the current task.
}
\label{fig:motivation}
\end{figure}

Existing methods mainly improve candidate retrieval or ranking.
Re-Invoke reformulates queries for zero-shot tool retrieval
\citep{chen2024reinvoke}, while ToolRerank improves candidate ordering through
adaptive reranking \citep{zheng2024toolrerank}.
Other approaches extend selection through generation or open-world function
calling \citep{wang2025toolgen,qin2025metatool}.
These methods provide strong mechanisms for proposing plausible candidates,
but relevance alone does not establish whether a candidate remains valid
after transfer.

Two additional constraints arise.
First, a relevant capability may lack an endpoint that jointly supports the
required operation, arguments, and input types; semantic relevance therefore
does not guarantee \emph{functional compatibility}.
Second, even a functionally valid capability may have been supported only
under particular source-side conditions, such as platform, locale, access
policy, or execution environment.
Its \emph{contextual applicability} must therefore be reassessed for the
current task.
As illustrated in Figure~\ref{fig:motivation}, zero-shot transfer must thus
consider relevance, functional compatibility, and contextual applicability.

Preliminary controlled diagnostics reveal both gaps, as shown in
Figure~\ref{fig:source_diagnostics}.
Among eligible tasks, $44.25\%$ contain functionally different
candidates with semantic scores less than $0.01$ apart,
indicating that semantic near-ties can mask functional incompatibility.
Holding the capability and callable function fixed, changing only
the task context reverses applicability in $15.20\%$ of determinate
pairs with an applicable base context ($62/408$).
These diagnostics motivate \emph{progressive validation}:
semantic relevance proposes candidates, endpoint evidence evaluates
functional compatibility, and task-conditioned source evidence
assesses contextual applicability.

Based on this observation, we propose \textbf{TACT}, an evidence-grounded
framework for zero-shot capability transfer.
Semantic Proposal provides a frozen relevance anchor.
Functional Validation scores task requirements against operation,
argument, and type evidence composed within individual endpoints.
Contextual Validation uses task-conditioned source evidence to
adjust relations whose support changes across contexts.
Finally, Validated Ranking combines these signals through bounded
gates, retaining the preceding estimate when a correction has no supporting
evidence.

We evaluate TACT under cross-domain and cross-dataset zero-shot transfer.
TACT achieves the strongest aggregate ranking performance, with larger gains
when semantically plausible candidates differ in functional compatibility or
contextual applicability.
On the cross-domain \textsc{Context} view, TACT improves NDCG@5 from
$0.518$ for the strongest baseline, PORTS, to $0.624$.
Ablation results confirm the contributions of both functional and contextual
validation, while downstream execution shows corresponding gains in task
completion across frozen agent backbones.

Our main contributions are:

\begin{itemize}
\item
We identify the \textbf{transfer-validity gap} in zero-shot capability
selection and distinguish semantic relevance, functional compatibility,
and contextual applicability through controlled diagnostics.

\item
We introduce TACT, which performs progressive transfer validation through
Semantic Proposal, Functional Validation, Contextual Validation, and
Validated Ranking.

\item
We evaluate TACT across cross-domain and cross-dataset transfer,
controlled candidate ambiguity, ablations, scaling, source-evidence
availability, and frozen-agent execution.
\end{itemize}

\section{Related Work}

\subsection{Tool Retrieval and Selection}

API-Bank, TaskBench, StableToolBench, and ToolRet provide benchmarks
for tool interaction, task automation, reproducible execution, and
retrieval, respectively
\cite{li2023apibank,shen2023taskbench,guo2024stabletoolbench,shi2025toolret}.
Within this setting, selection methods differ in the evidence used
to identify suitable candidates. ColBERTv2 captures fine-grained textual
matches \cite{santhanam2022colbertv2}; Re-Invoke and ToolDreamer enrich
retrieval with synthetic queries or hypothetical tools
\cite{chen2024reinvoke,sengupta2026tooldreamer}; and ToolRerank incorporates
tool hierarchy and unseen-tool adaptation \cite{zheng2024toolrerank}.
PORTS further introduces downstream model preferences
\cite{molfetta2025ports}, while task-aligned recommendation uses
historical bundles to adjust toolset composition
\cite{gao2025taskaligned}.
These signals improve candidate selection but do not explicitly score
operation, argument, and type evidence within individual callable
endpoints. \methodname{} incorporates this structure in Functional
Validation: requirement--endpoint matching aggregates evidence within
each endpoint before assessing support for the task.

\subsection{Transfer under Distribution Shift}

Transfer methods differ in how they share information across domains.
UniSRec learns reusable representations from item descriptions
\cite{hou2022unisrec}, whereas AREAD adapts expert sharing to individual
recommendation domains \cite{luo2025aread}. Domain-generalization
methods constrain what should remain stable: Fishr aligns gradient
variances across domains \cite{rame2022fishr}, non-stationary
generalization accounts for evolving domain patterns
\cite{pham2024nonstationary}, and sufficient invariant learning
preserves predictive invariant information \cite{kim2025sufficient}.
These approaches improve representation reuse or predictor robustness,
but leave unresolved whether source evidence for a functionally
compatible capability applies under the current task conditions.
\methodname{} therefore conditions source evidence on the current task
and limits the magnitude of a signed contextual correction. This
correction can increase or decrease a candidate's score; its bound
constrains the validation magnitude rather than imposing an upper bound
on the final score. When evidence is unavailable, the correction vanishes
and the gated functional estimate is retained.

\begin{figure}[t]
    \centering
    \includegraphics[
        width=\columnwidth
    ]{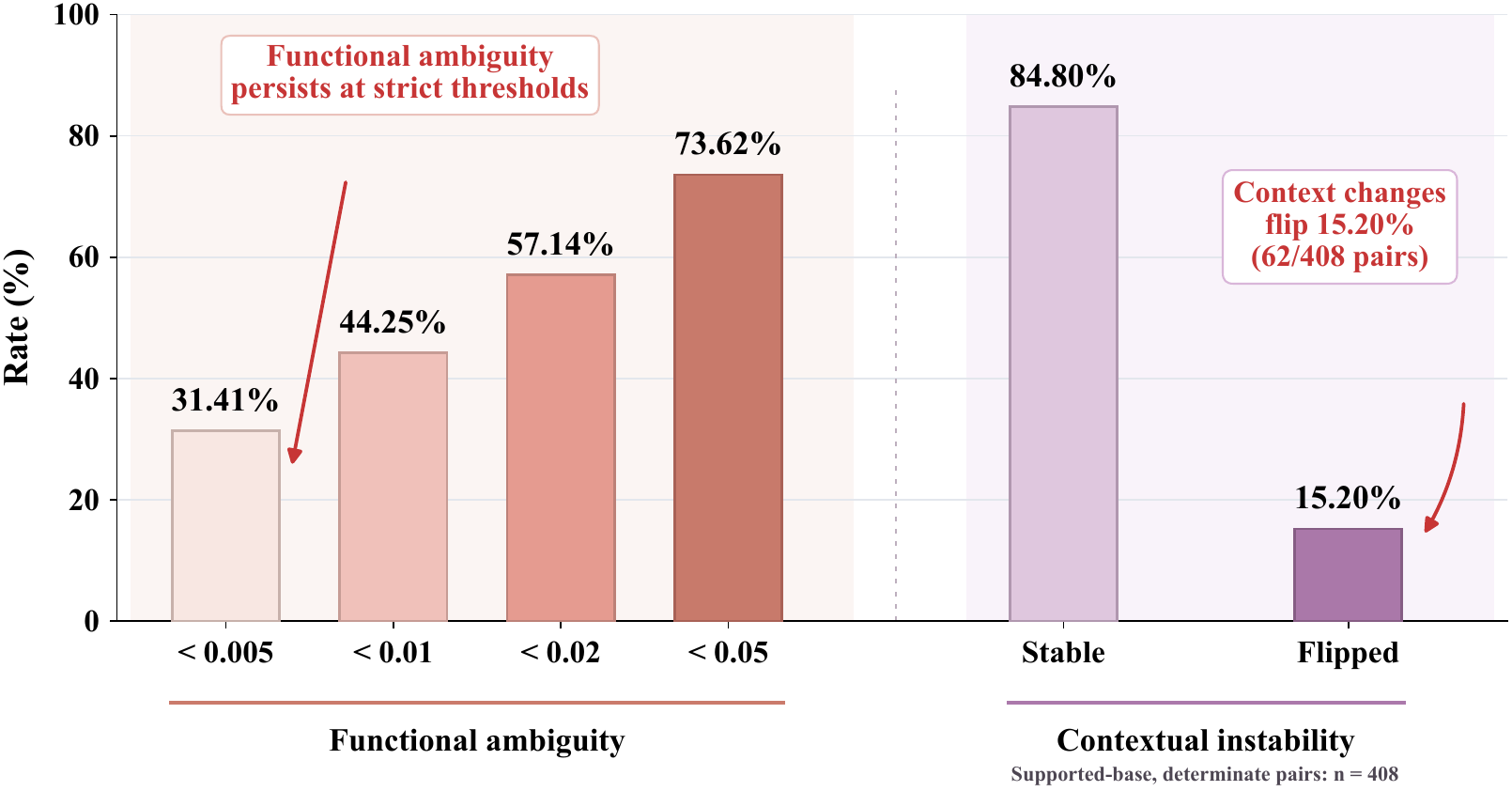}
    \caption{
    Controlled diagnostics of the transfer-validity gap.
    Left: functionally different candidates frequently remain close under
    frozen semantic relevance. Right: changing one task-context condition
    while preserving the capability and callable function flips applicability
    in $15.20\%$ of supported-base, determinate pairs ($62/408$).
    }
    \label{fig:source_diagnostics}
\end{figure}

\section{Methodology}
\label{sec:method}

\begin{figure*}[!t]
    \centering
    \includegraphics[width=1\textwidth]{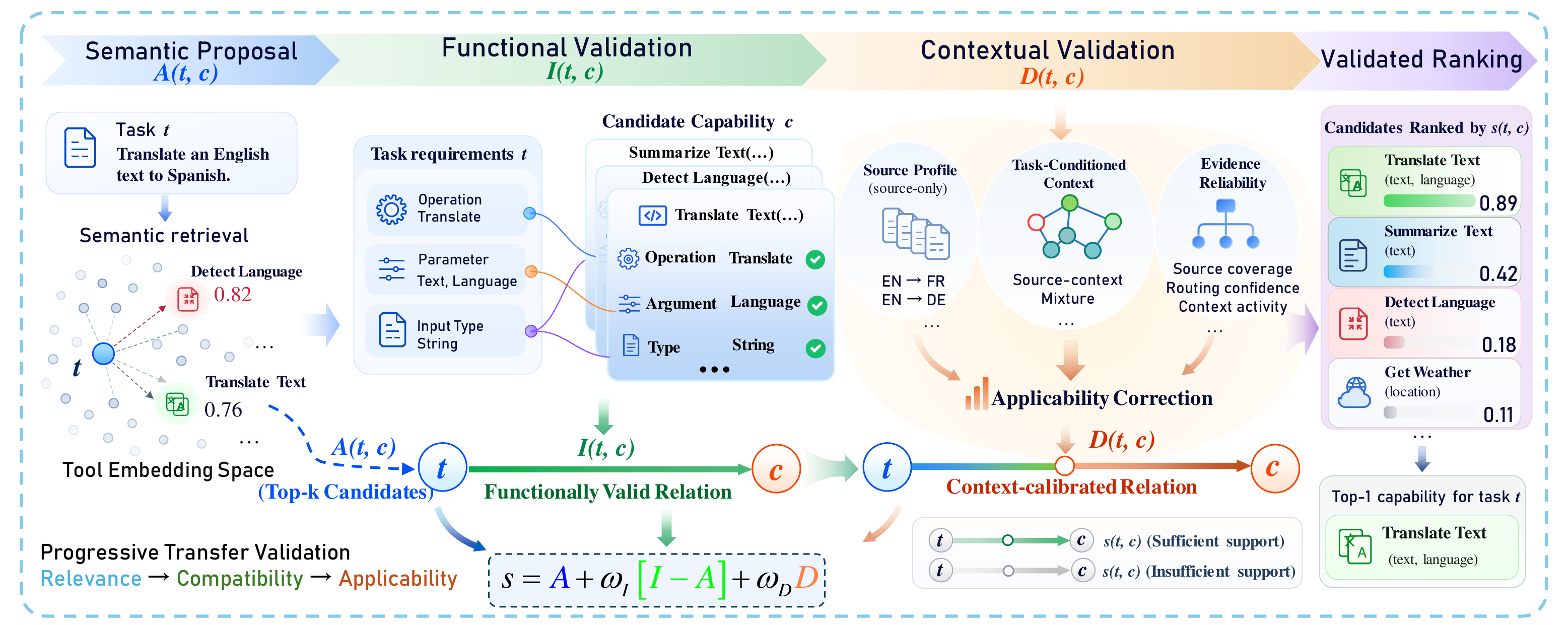}
    \caption{
    Overview of \methodname{}. Semantic Proposal supplies the anchor $A$;
    Functional Validation estimates Endpoint-Grounded compatibility $I$;
    Contextual Validation produces the signed applicability correction $D$;
    and Validated Ranking combines them through bounded gates. The semantic
    retrieval panel shows cosine similarities before the log-sigmoid
    calibration in Equation~\eqref{eq:semantic_anchor}; final-panel scores
    are schematic illustrations of the resulting order.}
    \label{fig:framework}
\end{figure*}

TACT follows a four-stage architecture: Semantic Proposal provides a
relevance anchor, Functional Validation evaluates endpoint compatibility,
Contextual Validation estimates task-conditioned applicability changes,
and Validated Ranking combines these signals through bounded gates,
as shown in Figure~\ref{fig:framework}.

\subsection{Problem Setup and Preliminary Diagnostics}
\label{sec:problem_formulation}

Given a task $t$ in a held-out context $d^\star$ and a candidate set
$\mathcal{C}_t$, our goal is to rank capabilities without using
target-side relevance labels for training, calibration, or model
selection. We define the source records available for transfer

\begin{equation}
\mathcal{S}_{-d^*}=\bigcup_{d\neq d^*}\mathcal{S}_d .
\label{eq:source_records}
\end{equation}

The model uses task and capability text, public endpoint schemas,
and source-derived statistics, but receives neither target-context
identity nor gold task decompositions.

We distinguish two transfer-validity gaps. Semantically similar candidates
may differ in endpoint support for required operations and inputs, separating
semantic relevance from functional compatibility. Functionally compatible
capabilities may change applicability across task conditions, separating
functional compatibility from contextual applicability.

Figure~\ref{fig:source_diagnostics} measures these gaps independently of
\methodname{}. For functional ambiguity, we consider tasks whose frozen
semantic Top-20 contains at least one clearly valid and one clearly invalid
candidate. For each eligible task, we compute
\[
d_t=
\min_{c^+\in\mathcal{V}_t,\;c^-\in\mathcal{I}_t}
\left|
s_{\mathrm{sem}}(t,c^+)-s_{\mathrm{sem}}(t,c^-)
\right|,
\]
where $\mathcal{V}_t$ and $\mathcal{I}_t$ denote functionally valid and
invalid candidates. Among 29,178 eligible tasks, $44.25\%$ contain a
functionally discordant pair with $d_t<0.01$, and $73.62\%$ do so with
$d_t<0.05$.

For contextual instability, we construct outcome-blind minimal pairs that
preserve the capability and callable function while changing one observed
task-context condition. After blinded review and adjudication, 408 of 500
pairs have supported base tasks and determinate applicability labels; 62
flip and 346 remain stable, yielding a $15.20\%$ flip rate.
These diagnostics motivate endpoint-grounded functional validation and
selective contextual validation. They are excluded from training, calibration,
and model selection; construction details and leakage controls appear in
Appendix~\ref{app:motivation_details}.

Primary ranking evaluation uses four fixed candidate views---Natural,
Semantic, Context, and Mixed---with a shared scorer, checkpoint, and calibration.
Candidate construction and evaluation details appear in
Appendices~\ref{app:construction_details}
and~\ref{app:evaluation_statistics}.

\subsection{Progressive Transfer Validation}
\label{sec:overview}

Recent tool-retrieval methods identify plausible candidates through semantic
retrieval or reranking \citep{chen2024reinvoke,zheng2024toolrerank}.
TACT begins with a semantic task--capability relation. Endpoint evidence
then assesses support for the required operations and inputs, while
task-conditioned source evidence adjusts applicability under the current
task conditions.

\paragraph{Validated Ranking.}
Let $A(t,c)$ denote the frozen semantic anchor, $I(t,c)$ the functional
compatibility score, and $D(t,c)$ the signed contextual correction.
The final score combines them as
\begin{equation}
\begin{aligned}
s(t,c)
={}&A(t,c)\\
&+\omega_I(t,c)\bigl[I(t,c)-A(t,c)\bigr]\\
&+\omega_D(t,c)D(t,c).
\end{aligned}
\label{eq:framework_score}
\end{equation}
Here $\omega_I,\omega_D\in[0,1]$ are selected using source contexts
only. The functional term interpolates between $A$ and $I$, while the
contextual term adds a gated signed correction. When contextual evidence
is unavailable, $D(t,c)=0$ and the score reduces to the functional path.

Training objectives, gate parameterization, and optimization are detailed in
Appendices~\ref{app:relational_objective}
--\ref{app:fusion_details}.

\subsection{Endpoint-Grounded Functional Validation}
\label{sec:functional}

Functional validation first refines the semantic retrieval proposal and then evaluates
its support in the candidate's documented endpoints.
A frozen encoder $E_0$ produces normalized task and capability
representations $e_t$ and $e_c$, from which we compute
\begin{equation}
\begin{aligned}
a(t,c)
&=
\frac{\langle e_t,e_c\rangle-\mu_a}{\tau_a},\\
A(t,c)
&=
\log\sigma\!\bigl(a(t,c)\bigr).
\end{aligned}
\label{eq:semantic_anchor}
\end{equation}
The parameters $\tau_a>0$ and $\mu_a$ are calibrated on source data.
The anchor preserves the cosine ranking and shares the log-sigmoid
scale of the functional score.

Source-trained relation encoders produce normalized states $z_t,z_c$
and a calibrated similarity $b(t,c)$. With
$\delta_r(t,c)=b(t,c)-a(t,c)$, we refine the semantic logit through
a bounded residual:
\begin{equation}
L_{\mathrm{rel}}(t,c)
=
a(t,c)
+
\alpha s_r
\tanh\!\left(
\frac{\delta_r(t,c)}{s_r}
\right).
\label{eq:relational_compatibility}
\end{equation}
Setting $\alpha=0$ recovers the semantic logit.
We then refine $L_{\mathrm{rel}}$ through requirement--endpoint matching,
combining operation, argument, and type evidence within each
documented endpoint.

\paragraph{Endpoint-coherent requirement matching.}
A deterministic task-only extractor produces a set of requirements
$\mathcal{N}_t=\{n_i\}_{i=1}^{M_t}$. For capability $c$, we retrieve a
set of $K_c\leq K$ deduplicated public endpoints shared across all
requirements. For requirement $i$ and endpoint $j$, let $o_{ij}$,
$a_{ij}$, and $y_{ij}$ denote operation, required-argument, and
argument-type evidence. Availability masks $m_j^a$ and $m_j^y$
indicate whether the corresponding schema fields are present.

Evidence is composed within each endpoint before matching, with missing
schema fields omitted from the weighted average:
\begin{equation}
v_{ij}
=
\frac{
o_{ij}
+\lambda_a m_j^a a_{ij}
+\lambda_y m_j^y y_{ij}
}{
1+\lambda_a m_j^a+\lambda_y m_j^y
}.
\label{eq:joint_endpoint_evidence}
\end{equation}

We search over partial one-to-one assignments
$X\in\{0,1\}^{M_t\times K_c}$, where each requirement is matched
to at most one endpoint and each endpoint is used at most once.
We denote the feasible set by $\mathcal{X}=\mathcal{X}(t,c)$.
Unmatched requirements use a shared null score $\nu$.
The resulting endpoint-set support is
\begin{equation}
\begin{aligned}
C_{\mathrm{ws}}(t,c)
={}&\frac{1}{M_t}
\max_{X\in\mathcal{X}}
\Bigl[
\sum_{i,j}X_{ij}v_{ij}\\
&\qquad\qquad
+\nu(M_t-|X|)
\Bigr].
\end{aligned}
\label{eq:witness_assignment}
\end{equation}
We calibrate $C_{\mathrm{ws}}(t,c)$ to $z_{\mathrm{ws}}(t,c)$ and compare
it with the relation logit. Let
$g_{\mathrm{ws}}=z_{\mathrm{ws}}-L_{\mathrm{rel}}$ and
$n_X=|X^\star|$ for a maximizing assignment. This discrepancy determines
the functional residual:
\begin{equation}
\Delta_I
=\rho_{\mathrm{ws}}
\begin{cases}
g_{\mathrm{ws}}, & n_X>0,\\
\min(g_{\mathrm{ws}},0), & n_X=0.
\end{cases}
\label{eq:functional_residual}
\end{equation}
The reliability factor $\rho_{\mathrm{ws}}$ scales the residual according
to the structural evidence, as defined in Appendix~\ref{app:witness_details}.
Missing requirements or endpoints give $\rho_{\mathrm{ws}}=0$.
When both are available but the assignment is all-null, we set
$\rho_{\mathrm{ws}}=1/2$ and retain a non-positive residual.
We incorporate the residual through a bounded update:
\begin{equation}
\Gamma_I(t,c)
=
\gamma_I s_I
\tanh\!\left(
\frac{\Delta_I(t,c)}{s_I}
\right).
\label{eq:functional_correction}
\end{equation}
The functional logit is
$L_I(t,c)=L_{\mathrm{rel}}(t,c)+\Gamma_I(t,c)$,
with $I(t,c)=\log\sigma(L_I(t,c))$.

During source-only training, semantic false friends provide an auxiliary
endpoint-set ranking objective that separates candidates remaining close
under semantic relevance. Reliability estimation, auxiliary supervision,
and exact assignment search are detailed in
Appendices~\ref{app:single_witness_details}
and~\ref{app:witness_details}.

\subsection{Evidence-Bounded Contextual Validation}
\label{sec:applicability}

Contextual validation uses source profiles to adjust the functional estimate
according to the conditions of the current task.

Positive source uses form a capability profile $p_c$, while the current
task induces a soft mixture $h_t$ over source-context centroids.
In cross-dataset transfer, source profiles are built from positive
Natural-view records and indexed by dataset--capability pairs. A target
capability obtains a profile by retrieving semantically similar source
capabilities within each source dataset and combining their profiles
using similarity and source-support weights. Semantic retrieval connects
source and target capabilities across different IDs and names.
The profile is set to zero when no source candidate exists or the maximum
similarity is below $0.25$.
A centered low-rank transformation of $(z_c,p_c,h_t)$ produces a
task-conditioned capability state $z_c^{\mathrm{ctx}}(t)$. We measure the
resulting relation change as
\begin{equation}
\Delta r(t,c)
=
\langle z_t,z_c^{\mathrm{ctx}}(t)\rangle
-
\langle z_t,z_c\rangle.
\label{eq:context_relation_change}
\end{equation}

\paragraph{Selective contextual validation.}
The applicability changes in Figure~\ref{fig:source_diagnostics} motivate
selective validation based on source confidence and observed task conditions.
Let $q_{\mathrm{conf}}(t,c)$ measure confidence in the available source
evidence and $q_{\mathrm{shift}}(t,c)$ measure task-observable context change:
\begin{equation}
\rho_{\mathrm{ctx}}(t,c)
=
q_{\mathrm{conf}}(t,c)
q_{\mathrm{shift}}(t,c).
\label{eq:context_reliability}
\end{equation}
This resembles selective prediction under distribution shift: uncertain
cases retain the default estimate \citep{liang2024selective}.

The contextual correction is
\begin{equation}
\begin{aligned}
D(t,c)
={}&
\gamma_D
\rho_{\mathrm{ctx}}(t,c)
g_{\mathrm{ev}}(t,c)\\
&\times
\tanh\!\left(
\frac{\Delta r(t,c)}{T_D}
\right),
\end{aligned}
\label{eq:contextual_correction}
\end{equation}
where $g_{\mathrm{ev}}\in[1-s_g,1+s_g]$ calibrates evidence reliability,
with $0\leq s_g\leq1$, $\gamma_D\geq0$, and $T_D>0$.
Because $\rho_{\mathrm{ctx}}\in[0,1]$, the correction satisfies
$|D(t,c)|\leq\gamma_D(1+s_g)$.
Its sign follows $\Delta r$, increasing or decreasing the score according
to the change in task--capability alignment.
All quantities use cross-fitted source evidence and task-observable
information.

A neutrality loss discourages corrections for context-neutral tasks:
\begin{equation}
\mathcal{L}_{\mathrm{neutral}}
=
\mathbb{E}\!\left[
(1-q_{\mathrm{shift}}(t,c))
D(t,c)^2
\right].
\label{eq:context_neutrality}
\end{equation}
The regularization weight is largest when $q_{\mathrm{shift}}$ is small
and decreases as evidence of context change increases.
Source confidence separately controls the correction through
$\rho_{\mathrm{ctx}}$; absent source evidence gives
$\rho_{\mathrm{ctx}}=0$ and hence $D=0$.

Source profiles, context routing, reliability features, and optimization
details are given in
Appendices~\ref{app:context_details}
--\ref{app:optimization_details}.

\subsection{Training and Model Selection}
\label{sec:training}

The two validation stages use supervision aligned with their respective
failure modes. Functional validation uses ordinary negatives and semantic
false friends, with the latter additionally supervising endpoint-set
matching. Contextual validation uses contextual counterfactuals together
with reliability and neutrality objectives that discourage unsupported
corrections.

All model strengths, checkpoints, and fusion parameters are selected using
source contexts only. After selection, the chosen configuration is
retrained on the complete source set and evaluated once on the held-out
target. For cross-dataset transfer, the scorer is unchanged; unlabeled
target embedding statistics are used only for label-free coordinate
alignment.

\section{Experiments}
\label{sec:experiments}

\begin{itemize}[
    labelindent=0pt,
    leftmargin=*,
    labelsep=0.5em,
    itemsep=1pt,
    parsep=0pt,
    topsep=2pt
]
    \item \textbf{RQ1:} Does TACT improve zero-shot capability ranking under
    semantic, contextual, and mixed transfer failures?

    \item \textbf{RQ2:} Do functional validation and contextual validation
    correct distinct transfer errors?

    \item \textbf{RQ3:} Do ranking gains improve execution completion
    across agent backbones?

    \item \textbf{RQ4:} Does TACT remain effective as capability libraries
    expand and source evidence becomes limited?
\end{itemize}

\subsection{Experimental Setup}
\label{sec:experimental_setup}

\paragraph{Transfer protocols.}
\textbf{Cross-domain} transfer uses leave-one-domain-out evaluation over
49 ToolBench domains, with the other 48 providing source supervision.
\textbf{Cross-dataset} transfer holds out each of
ToolBench~\citep{qin2023toolllm},
TaskBench~\citep{shen2023taskbench},
MetaTool~\citep{huang2024metatool}, and
APIBench~\citep{patil2024gorilla},
using the other three as sources.
These protocols assess within-benchmark generalization and broader shifts
in task and capability distributions, respectively.
Training, calibration, and model selection use source data only;
cross-dataset transfer also uses unlabeled target embeddings
for label-free coordinate alignment.
Results are averaged over seeds 42, 43, and 44, then equally
across scored targets.

\paragraph{Baselines and metrics.}
Baselines include frozen retrieval (Random and Semantic Cosine),
source-trained Pointwise MLP;
transfer methods ZESRec~\citep{ding2021zesrec},
UniSRec~\citep{hou2022unisrec}, and AREAD~\citep{luo2025aread};
and tool-retrieval methods Re-Invoke~\citep{chen2024reinvoke},
PORTS~\citep{molfetta2025ports}, and
ToolDreamer~\citep{sengupta2026tooldreamer}.
Endpoint Schema Match uses public endpoint and schema evidence,
while MultiPool pools source data.
Re-Invoke, ToolDreamer, and PORTS retain their original configurations
under the shared ranking interface; PORTS uses the authors' code.
Backbones, encoders, and source-data use appear in
Appendix~\ref{app:baseline_details}.
We report MRR, NDCG@5, and HR@3; split statistics and implementation
details appear in Appendices~\ref{app:evaluation_statistics}
and~\ref{app:implementation_details}.

\begin{table*}[!t]
\centering\small
\setlength{\tabcolsep}{1.8pt}
\renewcommand{\arraystretch}{0.93}
\resizebox{\textwidth}{!}{%
\begin{tabular}{@{}>{\centering\arraybackslash}m{1.2cm}>{\raggedright\arraybackslash}m{2.35cm}*{3}{>{\centering\arraybackslash}m{.97cm}}@{\hspace{2pt}{\color{tactViewSep}\vrule width .3pt}\hspace{2pt}}*{3}{>{\centering\arraybackslash}m{.97cm}}@{\hspace{2pt}{\color{tactViewSep}\vrule width .3pt}\hspace{2pt}}*{3}{>{\centering\arraybackslash}m{.97cm}}@{\hspace{2pt}{\color{tactViewSep}\vrule width .3pt}\hspace{2pt}}*{3}{>{\centering\arraybackslash}m{.97cm}}@{}}
\toprule
Protocol & Model & \multicolumn{3}{c}{Natural} & \multicolumn{3}{c}{Semantic} & \multicolumn{3}{c}{Context} & \multicolumn{3}{c}{Mixed} \\
\cmidrule(lr){3-5}\cmidrule(lr){6-8}\cmidrule(lr){9-11}\cmidrule(lr){12-14}
& & MRR & N@5 & H@3 & MRR & N@5 & H@3 & MRR & N@5 & H@3 & MRR & N@5 & H@3 \\
\midrule
\cellcolor{tactProtocolFill} & \multirow{2}{*}{Random} & 0.227 & 0.157 & 0.220 & 0.222 & 0.157 & 0.209 & 0.222 & 0.154 & 0.212 & 0.219 & 0.153 & 0.205 \\[-.4ex]
\cellcolor{tactProtocolFill} & & \tactstd{0.001} & \tactstd{0.001} & \tactstd{0.003} & \tactstd{0.004} & \tactstd{0.006} & \tactstd{0.008} & \tactstd{0.001} & \tactstd{0.001} & \tactstd{0.004} & \tactstd{0.001} & \tactstd{0.003} & \tactstd{0.004} \\[.22ex]
\cellcolor{tactProtocolFill} & \multirow{2}{*}{Semantic Cosine} & \underline{0.881} & 0.840 & \underline{0.911} & 0.263 & 0.246 & 0.305 & 0.259 & 0.225 & 0.297 & 0.213 & 0.165 & 0.190 \\[-.4ex]
\cellcolor{tactProtocolFill} & & \tactstd{0.000} & \tactstd{0.000} & \tactstd{0.000} & \tactstd{0.000} & \tactstd{0.000} & \tactstd{0.000} & \tactstd{0.000} & \tactstd{0.000} & \tactstd{0.000} & \tactstd{0.000} & \tactstd{0.000} & \tactstd{0.000} \\[.22ex]
\cellcolor{tactProtocolFill} & \multirow{2}{*}{Pointwise MLP} & 0.740 & 0.699 & 0.823 & 0.440 & 0.410 & 0.500 & 0.435 & 0.392 & 0.487 & 0.399 & 0.355 & 0.442 \\[-.4ex]
\cellcolor{tactProtocolFill} & & \tactstd{0.002} & \tactstd{0.002} & \tactstd{0.003} & \tactstd{0.004} & \tactstd{0.004} & \tactstd{0.008} & \tactstd{0.006} & \tactstd{0.007} & \tactstd{0.011} & \tactstd{0.008} & \tactstd{0.007} & \tactstd{0.012} \\[.22ex]
\cellcolor{tactProtocolFill} & \multirow{2}{*}{ZESRec} & 0.210 & 0.139 & 0.169 & 0.215 & 0.146 & 0.173 & 0.209 & 0.138 & 0.165 & 0.206 & 0.133 & 0.161 \\[-.4ex]
\cellcolor{tactProtocolFill} & & \tactstd{0.004} & \tactstd{0.003} & \tactstd{0.005} & \tactstd{0.008} & \tactstd{0.009} & \tactstd{0.009} & \tactstd{0.002} & \tactstd{0.002} & \tactstd{0.002} & \tactstd{0.003} & \tactstd{0.004} & \tactstd{0.003} \\[.22ex]
\cellcolor{tactProtocolFill} & \multirow{2}{*}{UniSRec} & 0.671 & 0.634 & 0.772 & 0.468 & 0.439 & 0.546 & 0.472 & 0.430 & 0.541 & 0.421 & 0.380 & 0.480 \\[-.4ex]
\cellcolor{tactProtocolFill} & & \tactstd{0.001} & \tactstd{0.001} & \tactstd{0.001} & \tactstd{0.004} & \tactstd{0.004} & \tactstd{0.005} & \tactstd{0.001} & \tactstd{0.001} & \tactstd{0.004} & \tactstd{0.004} & \tactstd{0.004} & \tactstd{0.007} \\[.22ex]
\cellcolor{tactProtocolFill} & \multirow{2}{*}{AREAD} & 0.440 & 0.360 & 0.455 & 0.322 & 0.255 & 0.318 & 0.303 & 0.236 & 0.297 & 0.287 & 0.220 & 0.274 \\[-.4ex]
\cellcolor{tactProtocolFill} & & \tactstd{0.005} & \tactstd{0.006} & \tactstd{0.007} & \tactstd{0.007} & \tactstd{0.007} & \tactstd{0.008} & \tactstd{0.008} & \tactstd{0.008} & \tactstd{0.008} & \tactstd{0.008} & \tactstd{0.007} & \tactstd{0.008} \\[.22ex]
\cellcolor{tactProtocolFill} & \multirow{2}{*}{Endpoint Schema} & 0.740 & 0.655 & 0.777 & 0.591 & \underline{0.547} & \underline{0.661} & 0.527 & 0.473 & 0.588 & 0.505 & 0.467 & \underline{0.587} \\[-.4ex]
\cellcolor{tactProtocolFill} & & \tactstd{0.000} & \tactstd{0.000} & \tactstd{0.000} & \tactstd{0.000} & \tactstd{0.000} & \tactstd{0.000} & \tactstd{0.000} & \tactstd{0.000} & \tactstd{0.000} & \tactstd{0.000} & \tactstd{0.000} & \tactstd{0.000} \\[.22ex]
\cellcolor{tactProtocolFill} & \multirow{2}{*}{Re-Invoke} & 0.744 & 0.674 & 0.789 & 0.306 & 0.267 & 0.342 & 0.311 & 0.257 & 0.331 & 0.271 & 0.216 & 0.273 \\[-.4ex]
\cellcolor{tactProtocolFill} & & \tactstd{0.006} & \tactstd{0.003} & \tactstd{0.004} & \tactstd{0.007} & \tactstd{0.002} & \tactstd{0.003} & \tactstd{0.006} & \tactstd{0.004} & \tactstd{0.005} & \tactstd{0.006} & \tactstd{0.007} & \tactstd{0.002} \\[.22ex]
\cellcolor{tactProtocolFill} & \multirow{2}{*}{PORTS} & 0.741 & 0.653 & 0.752 & \underline{0.597} & 0.546 & 0.659 & \underline{0.575} & \underline{0.518} & \underline{0.623} & \underline{0.520} & \underline{0.471} & 0.579 \\[-.4ex]
\cellcolor{tactProtocolFill} & & \tactstd{0.003} & \tactstd{0.009} & \tactstd{0.003} & \tactstd{0.005} & \tactstd{0.007} & \tactstd{0.001} & \tactstd{0.007} & \tactstd{0.004} & \tactstd{0.008} & \tactstd{0.005} & \tactstd{0.006} & \tactstd{0.001} \\[.22ex]
\cellcolor{tactProtocolFill} & \multirow{2}{*}{ToolDreamer} & 0.879 & \underline{0.841} & 0.910 & 0.296 & 0.284 & 0.358 & 0.293 & 0.256 & 0.324 & 0.244 & 0.199 & 0.244 \\[-.4ex]
\cellcolor{tactProtocolFill} & & \tactstd{0.003} & \tactstd{0.003} & \tactstd{0.006} & \tactstd{0.007} & \tactstd{0.003} & \tactstd{0.007} & \tactstd{0.007} & \tactstd{0.003} & \tactstd{0.007} & \tactstd{0.007} & \tactstd{0.002} & \tactstd{0.002} \\[.22ex]
\cellcolor{tactProtocolFill} & \multirow{2}{*}{MultiPool} & 0.812 & 0.761 & 0.875 & 0.406 & 0.382 & 0.490 & 0.395 & 0.355 & 0.467 & 0.339 & 0.298 & 0.389 \\[-.4ex]
\cellcolor{tactProtocolFill} & & \tactstd{0.007} & \tactstd{0.008} & \tactstd{0.006} & \tactstd{0.006} & \tactstd{0.007} & \tactstd{0.004} & \tactstd{0.007} & \tactstd{0.005} & \tactstd{0.004} & \tactstd{0.006} & \tactstd{0.007} & \tactstd{0.006} \\[.22ex]
\cmidrule(lr){2-14}
\rowcolor{tactRowFill}
\cellcolor{tactProtocolFill} & \textbf{TACT (Full)} & \textbf{0.915} & \textbf{0.873} & \textbf{0.952} & \textbf{0.629} & \textbf{0.604} & \textbf{0.726} & \textbf{0.672} & \textbf{0.624} & \textbf{0.755} & \textbf{0.564} & \textbf{0.529} & \textbf{0.644} \\[-.4ex]
\rowcolor{tactRowFill}
\cellcolor{tactProtocolFill} & & \tactstd{0.004} & \tactstd{0.005} & \tactstd{0.003} & \tactstd{0.005} & \tactstd{0.005} & \tactstd{0.007} & \tactstd{0.003} & \tactstd{0.003} & \tactstd{0.004} & \tactstd{0.003} & \tactstd{0.006} & \tactstd{0.005} \\[.22ex]
\rowcolor{tactGainFill}\cellcolor{tactProtocolFill}\multirow{-25}{*}{\shortstack{Cross-\\domain}} & {\scriptsize\emph{$\Delta$ vs.\ best (pp)}} & +3.4 & +3.2 & +4.1 & +3.2 & +5.7 & +6.5 & +9.7 & +10.6 & +13.2 & +4.4 & +5.8 & +5.7 \\
\midrule
\cellcolor{tactProtocolFill} & \multirow{2}{*}{Random} & 0.314 & 0.274 & 0.343 & 0.311 & 0.273 & 0.341 & 0.298 & 0.249 & 0.316 & 0.298 & 0.250 & 0.316 \\[-.4ex]
\cellcolor{tactProtocolFill} & & \tactstd{0.000} & \tactstd{0.001} & \tactstd{0.001} & \tactstd{0.001} & \tactstd{0.001} & \tactstd{0.001} & \tactstd{0.001} & \tactstd{0.001} & \tactstd{0.002} & \tactstd{0.001} & \tactstd{0.001} & \tactstd{0.001} \\[.22ex]
\cellcolor{tactProtocolFill} & \multirow{2}{*}{Semantic Cosine} & 0.734 & 0.708 & 0.810 & 0.565 & 0.528 & 0.621 & 0.516 & 0.477 & 0.545 & 0.475 & 0.427 & 0.482 \\[-.4ex]
\cellcolor{tactProtocolFill} & & \tactstd{0.000} & \tactstd{0.000} & \tactstd{0.000} & \tactstd{0.000} & \tactstd{0.000} & \tactstd{0.000} & \tactstd{0.000} & \tactstd{0.000} & \tactstd{0.000} & \tactstd{0.000} & \tactstd{0.000} & \tactstd{0.000} \\[.22ex]
\cellcolor{tactProtocolFill} & \multirow{2}{*}{Pointwise MLP} & 0.545 & 0.519 & 0.630 & 0.456 & 0.430 & 0.535 & 0.458 & 0.425 & 0.518 & 0.437 & 0.403 & 0.498 \\[-.4ex]
\cellcolor{tactProtocolFill} & & \tactstd{0.002} & \tactstd{0.005} & \tactstd{0.006} & \tactstd{0.003} & \tactstd{0.004} & \tactstd{0.006} & \tactstd{0.008} & \tactstd{0.008} & \tactstd{0.008} & \tactstd{0.005} & \tactstd{0.004} & \tactstd{0.004} \\[.22ex]
\cellcolor{tactProtocolFill} & \multirow{2}{*}{ZESRec} & 0.351 & 0.317 & 0.391 & 0.339 & 0.304 & 0.377 & 0.359 & 0.319 & 0.399 & 0.330 & 0.290 & 0.370 \\[-.4ex]
\cellcolor{tactProtocolFill} & & \tactstd{0.007} & \tactstd{0.008} & \tactstd{0.012} & \tactstd{0.009} & \tactstd{0.010} & \tactstd{0.012} & \tactstd{0.002} & \tactstd{0.004} & \tactstd{0.007} & \tactstd{0.001} & \tactstd{0.005} & \tactstd{0.009} \\[.22ex]
\cellcolor{tactProtocolFill} & \multirow{2}{*}{UniSRec} & 0.429 & 0.407 & 0.506 & 0.392 & 0.366 & 0.458 & 0.407 & 0.374 & 0.461 & 0.377 & 0.344 & 0.434 \\[-.4ex]
\cellcolor{tactProtocolFill} & & \tactstd{0.001} & \tactstd{0.001} & \tactstd{0.005} & \tactstd{0.001} & \tactstd{0.002} & \tactstd{0.004} & \tactstd{0.005} & \tactstd{0.006} & \tactstd{0.008} & \tactstd{0.001} & \tactstd{0.002} & \tactstd{0.003} \\[.22ex]
\cellcolor{tactProtocolFill} & \multirow{2}{*}{AREAD} & 0.454 & 0.426 & 0.533 & 0.405 & 0.375 & 0.476 & 0.408 & 0.374 & 0.464 & 0.387 & 0.354 & 0.450 \\[-.4ex]
\cellcolor{tactProtocolFill} & & \tactstd{0.001} & \tactstd{0.001} & \tactstd{0.001} & \tactstd{0.001} & \tactstd{0.003} & \tactstd{0.009} & \tactstd{0.003} & \tactstd{0.004} & \tactstd{0.006} & \tactstd{0.001} & \tactstd{0.002} & \tactstd{0.004} \\[.22ex]
\cellcolor{tactProtocolFill} & \multirow{2}{*}{Endpoint Schema} & 0.522 & 0.477 & 0.568 & 0.512 & 0.480 & 0.573 & 0.470 & 0.432 & 0.514 & 0.451 & 0.410 & 0.501 \\[-.4ex]
\cellcolor{tactProtocolFill} & & \tactstd{0.000} & \tactstd{0.000} & \tactstd{0.000} & \tactstd{0.000} & \tactstd{0.000} & \tactstd{0.000} & \tactstd{0.000} & \tactstd{0.000} & \tactstd{0.000} & \tactstd{0.000} & \tactstd{0.000} & \tactstd{0.000} \\[.22ex]
\cellcolor{tactProtocolFill} & \multirow{2}{*}{Re-Invoke} & 0.643 & 0.594 & 0.719 & 0.522 & 0.463 & 0.570 & 0.476 & 0.415 & 0.502 & 0.457 & 0.397 & 0.482 \\[-.4ex]
\cellcolor{tactProtocolFill} & & \tactstd{0.003} & \tactstd{0.006} & \tactstd{0.004} & \tactstd{0.008} & \tactstd{0.006} & \tactstd{0.002} & \tactstd{0.006} & \tactstd{0.003} & \tactstd{0.007} & \tactstd{0.005} & \tactstd{0.008} & \tactstd{0.006} \\[.22ex]
\cellcolor{tactProtocolFill} & \multirow{2}{*}{PORTS} & 0.755 & 0.730 & 0.827 & 0.581 & 0.543 & 0.624 & 0.534 & 0.506 & 0.578 & 0.489 & 0.453 & 0.514 \\[-.4ex]
\cellcolor{tactProtocolFill} & & \tactstd{0.009} & \tactstd{0.010} & \tactstd{0.011} & \tactstd{0.010} & \tactstd{0.011} & \tactstd{0.012} & \tactstd{0.009} & \tactstd{0.011} & \tactstd{0.014} & \tactstd{0.008} & \tactstd{0.010} & \tactstd{0.011} \\[.22ex]
\cellcolor{tactProtocolFill} & \multirow{2}{*}{ToolDreamer} & 0.745 & 0.725 & 0.826 & 0.580 & 0.546 & \underline{0.640} & \underline{0.544} & \underline{0.510} & \underline{0.583} & \underline{0.500} & \underline{0.463} & \underline{0.530} \\[-.4ex]
\cellcolor{tactProtocolFill} & & \tactstd{0.007} & \tactstd{0.005} & \tactstd{0.005} & \tactstd{0.007} & \tactstd{0.008} & \tactstd{0.007} & \tactstd{0.005} & \tactstd{0.007} & \tactstd{0.003} & \tactstd{0.004} & \tactstd{0.007} & \tactstd{0.008} \\[.22ex]
\cellcolor{tactProtocolFill} & \multirow{2}{*}{MultiPool} & \underline{0.756} & \underline{0.733} & \underline{0.830} & \underline{0.583} & \underline{0.547} & 0.639 & 0.537 & 0.500 & 0.500 & 0.499 & 0.444 & 0.519 \\[-.4ex]
\cellcolor{tactProtocolFill} & & \tactstd{0.006} & \tactstd{0.007} & \tactstd{0.007} & \tactstd{0.005} & \tactstd{0.004} & \tactstd{0.009} & \tactstd{0.006} & \tactstd{0.004} & \tactstd{0.006} & \tactstd{0.008} & \tactstd{0.006} & \tactstd{0.005} \\[.22ex]
\cmidrule(lr){2-14}
\rowcolor{tactRowFill}
\cellcolor{tactProtocolFill} & \textbf{TACT (Full)} & \textbf{0.763} & \textbf{0.746} & \textbf{0.836} & \textbf{0.613} & \textbf{0.580} & \textbf{0.668} & \textbf{0.574} & \textbf{0.543} & \textbf{0.613} & \textbf{0.538} & \textbf{0.500} & \textbf{0.572} \\[-.4ex]
\rowcolor{tactRowFill}
\cellcolor{tactProtocolFill} & & \tactstd{0.002} & \tactstd{0.003} & \tactstd{0.004} & \tactstd{0.003} & \tactstd{0.003} & \tactstd{0.003} & \tactstd{0.002} & \tactstd{0.002} & \tactstd{0.002} & \tactstd{0.004} & \tactstd{0.004} & \tactstd{0.002} \\[.22ex]
\rowcolor{tactGainFill}\cellcolor{tactProtocolFill}\multirow{-25}{*}{\shortstack{Cross-\\dataset}} & {\scriptsize\emph{$\Delta$ vs.\ best (pp)}} & +0.7 & +1.3 & +0.6 & +3.0 & +3.3 & +2.8 & +3.0 & +3.3 & +3.0 & +3.8 & +3.7 & +4.2 \\
\bottomrule
\end{tabular}%
}
\caption{
\textbf{Overall performance} on zero-shot capability ranking under cross-domain
and cross-dataset transfer across four candidate views, reported as
mean $\pm$ standard deviation over three runs.
Best and second-best means are bolded and underlined, respectively.
The final row of each protocol reports TACT's absolute gain over
the strongest non-TACT baseline in percentage points (pp).
}
\label{tab:main_results}
\end{table*}
\subsection{Overall Performance (RQ1)}
\label{sec:overall_results}

TACT achieves the highest aggregate scores across both transfer protocols,
with larger NDCG@5 gains on the constructed views than on Natural.
Table~\ref{tab:main_results} shows the largest cross-domain improvement
on Context, where NDCG@5 rises from 0.518 for PORTS to 0.624 for TACT.
On Natural, where semantic retrieval is already strong, TACT improves
NDCG@5 from 0.841 for ToolDreamer to 0.873.

The same pattern holds in cross-dataset transfer.
On Mixed, TACT increases NDCG@5 from 0.463 for ToolDreamer to 0.500,
compared with an increase from 0.733 for MultiPool to 0.746 on Natural.
The larger improvement on Mixed indicates that validation is especially
useful when the candidate set combines functional and contextual ambiguity.

TACT also improves on retrieval using the same public schema evidence.
On cross-domain Semantic, it achieves NDCG@5 of 0.604, compared with
0.547 for Endpoint Schema Match, the strongest baseline on this metric.
This comparison supports combining endpoint evidence with learned
task--capability relations and contextual validation.
Per-target Full-model results and held-out-group comparisons appear
in Appendices~\ref{app:full_per_target} and~\ref{app:group_held_out}.

\subsection{Validation-Stage Analysis (RQ2)}
\label{sec:ablation}
\begin{table*}[t]
\centering
\small
\setlength{\tabcolsep}{1.8pt}
\renewcommand{\arraystretch}{1.08}

\resizebox{\textwidth}{!}{%
\begin{tabular}{
@{}
>{\centering\arraybackslash}m{1.35cm}
>{\raggedright\arraybackslash}m{2.55cm}
*{12}{>{\centering\arraybackslash}m{1.04cm}}
@{}
}

\toprule

\multirow{2}{*}{Protocol}
& \multirow{2}{*}{Variant}
& \multicolumn{3}{c}{Natural}
& \multicolumn{3}{c}{Semantic}
& \multicolumn{3}{c}{Context}
& \multicolumn{3}{c}{Mixed} \\

\cmidrule(lr){3-5}
\cmidrule(lr){6-8}
\cmidrule(lr){9-11}
\cmidrule(lr){12-14}

&
& MRR & N@5 & H@3
& MRR & N@5 & H@3
& MRR & N@5 & H@3
& MRR & N@5 & H@3 \\

\midrule

\rowcolor{tactrow}
\multirow[c]{4}{1.35cm}{\centering\shortstack{Cross\\domain}}
& \textbf{\methodname{} (Full)}
& \textbf{0.915} & \textbf{0.873} & \textbf{0.952}
& \textbf{0.629} & \textbf{0.604} & \textbf{0.726}
& \textbf{0.672} & \textbf{0.624} & \textbf{0.755}
& \textbf{0.564} & \textbf{0.529} & \textbf{0.644} \\

& w/o Context
& 0.899 & 0.853 & 0.939
& 0.620 & 0.592 & 0.720
& 0.594 & 0.552 & 0.677
& 0.521 & 0.483 & 0.594 \\

& w/o Functional
& 0.903 & 0.859 & 0.941
& 0.568 & 0.545 & 0.660
& 0.605 & 0.557 & 0.683
& 0.504 & 0.467 & 0.579 \\

& Anchor only
& 0.881 & 0.840 & 0.911
& 0.263 & 0.246 & 0.305
& 0.259 & 0.225 & 0.297
& 0.213 & 0.165 & 0.190 \\

\midrule

\rowcolor{tactrow}
\multirow[c]{4}{1.35cm}{\centering\shortstack{Cross\\dataset}}
& \textbf{\methodname{} (Full)}
& \textbf{0.763} & \textbf{0.746} & \textbf{0.836}
& \textbf{0.613} & \textbf{0.580} & \textbf{0.668}
& \textbf{0.574} & \textbf{0.543} & \textbf{0.613}
& \textbf{0.538} & \textbf{0.500} & \textbf{0.572} \\

& w/o Context
& 0.749 & 0.730 & 0.818
& 0.611 & 0.578 & 0.667
& 0.568 & 0.534 & 0.602
& 0.535 & 0.497 & 0.565 \\

& w/o Functional
& 0.755 & 0.730 & 0.829
& 0.580 & 0.546 & 0.636
& 0.538 & 0.500 & 0.567
& 0.497 & 0.452 & 0.513 \\

& Anchor only
& 0.734 & 0.708 & 0.810
& 0.565 & 0.528 & 0.621
& 0.516 & 0.477 & 0.545
& 0.475 & 0.427 & 0.482 \\

\bottomrule
\end{tabular}%
}

\caption{
\textbf{Ablation Study.}
Cross-domain and cross-dataset transfer without Functional or
Contextual Validation; \emph{Anchor only} uses the frozen semantic path.
Trainable variants are independently fitted and selected on source data,
unlike RQ4's fixed-checkpoint diagnostic. Best results per setting are bold.
}
\label{tab:compact_branch_ablation}
\end{table*}

The full model performs best across both protocols, with functional
validation contributing broadly and contextual validation having a more
selective effect. Table~\ref{tab:compact_branch_ablation} compares variants
trained and selected independently using source data.

Removing functional validation reduces cross-domain NDCG@5 by 0.059
on Semantic, 0.067 on Context, and 0.062 on Mixed.
Its contribution persists across datasets, where removal lowers
NDCG@5 by 0.048 on Mixed. Contextual validation contributes most on cross-domain Context and Mixed.
Removing it lowers NDCG@5 by 0.072 and 0.046, compared with 0.020
on Natural and 0.012 on Semantic.
Its effect is smaller in cross-dataset transfer, where the reductions
are 0.016 on Natural, 0.002 on Semantic, 0.009 on Context, and 0.003
on Mixed. The contextual contribution therefore depends more strongly
on the transfer setting.

\begin{figure*}[t]
    \centering
    \includegraphics[
        width=\textwidth
    ]{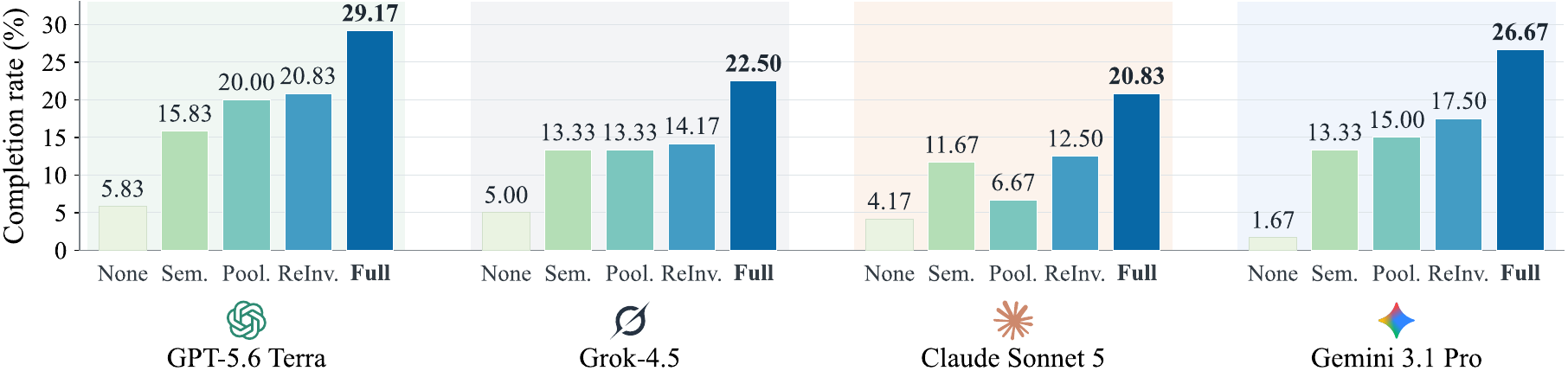}
    \caption{
\textbf{Downstream Execution.}
Completion rates on 120 solvable StableToolBench queries
under cached execution with four frozen agent backbones.
None denotes no tools; Sem., Pool., ReInv., and Full use
Top-5 capabilities from Semantic Cosine, MultiPool,
Re-Invoke, and TACT, respectively.
}
    \label{fig:downstream_execution}
\end{figure*}

\subsection{Downstream Cached Execution (RQ3)}
\label{sec:execution_results}

\methodname{} achieves the highest task completion rate across all four
frozen agent backbones. Figure~\ref{fig:downstream_execution} reports
results on 120 solvable StableToolBench queries under cached execution.

Compared with Re-Invoke, the strongest competing selector, completion
increases from 20.83\% to 29.17\% on GPT-5.6 Terra,
from 14.17\% to 22.50\% on Grok-4.5,
from 12.50\% to 20.83\% on Claude Sonnet 5,
and from 17.50\% to 26.67\% on Gemini 3.1 Pro. All retrieval policies select the top five capabilities from the same
3,442-capability library and use identical queries and execution budgets.
Ranking configurations and agent policies remain fixed throughout
evaluation. Under these shared conditions, the completion gains show
that improved capability selection benefits downstream execution
across different agent models.
\begin{table}[t]
\centering
\small
\setlength{\tabcolsep}{3.0pt}
\renewcommand{\arraystretch}{1.15}

\resizebox{\columnwidth}{!}{%
\begin{tabular}{l*{4}{cc}}
\toprule
\multirow{2}{*}{\textbf{Scale}}
& \multicolumn{2}{c}{\textbf{Natural}}
& \multicolumn{2}{c}{\textbf{Semantic}}
& \multicolumn{2}{c}{\textbf{Context}}
& \multicolumn{2}{c}{\textbf{Mixed}} \\

\cmidrule(lr){2-3}
\cmidrule(lr){4-5}
\cmidrule(lr){6-7}
\cmidrule(lr){8-9}

& Sem. & \methodname{}
& Sem. & \methodname{}
& Sem. & \methodname{}
& Sem. & \methodname{} \\

\midrule
Base
& 0.8397 & \textbf{0.8725}
& 0.2458 & \textbf{0.6035}
& 0.2246 & \textbf{0.6244}
& 0.1654 & \textbf{0.5288} \\

100
& 0.4319 & \textbf{0.4454}
& 0.0406 & \textbf{0.2930}
& 0.0780 & \textbf{0.3242}
& 0.0600 & \textbf{0.2862} \\

500
& 0.4015 & \textbf{0.4100}
& 0.0346 & \textbf{0.2478}
& 0.0626 & \textbf{0.2782}
& 0.0508 & \textbf{0.2462} \\

Full
& 0.3822 & \textbf{0.4011}
& 0.0333 & \textbf{0.2315}
& 0.0563 & \textbf{0.2602}
& 0.0483 & \textbf{0.2311} \\

\bottomrule
\end{tabular}%
}

\caption{
\textbf{Library Scaling.}
Cross-domain NDCG@5 by candidate-pool size.
Base: evaluation manifest; Full: 3,442 capabilities;
Sem.: Semantic Cosine. Pairwise best in bold.
}
\label{tab:candidate_scaling}
\end{table}

\begin{figure}[t]
    \centering
    \includegraphics[
        width=\columnwidth
    ]{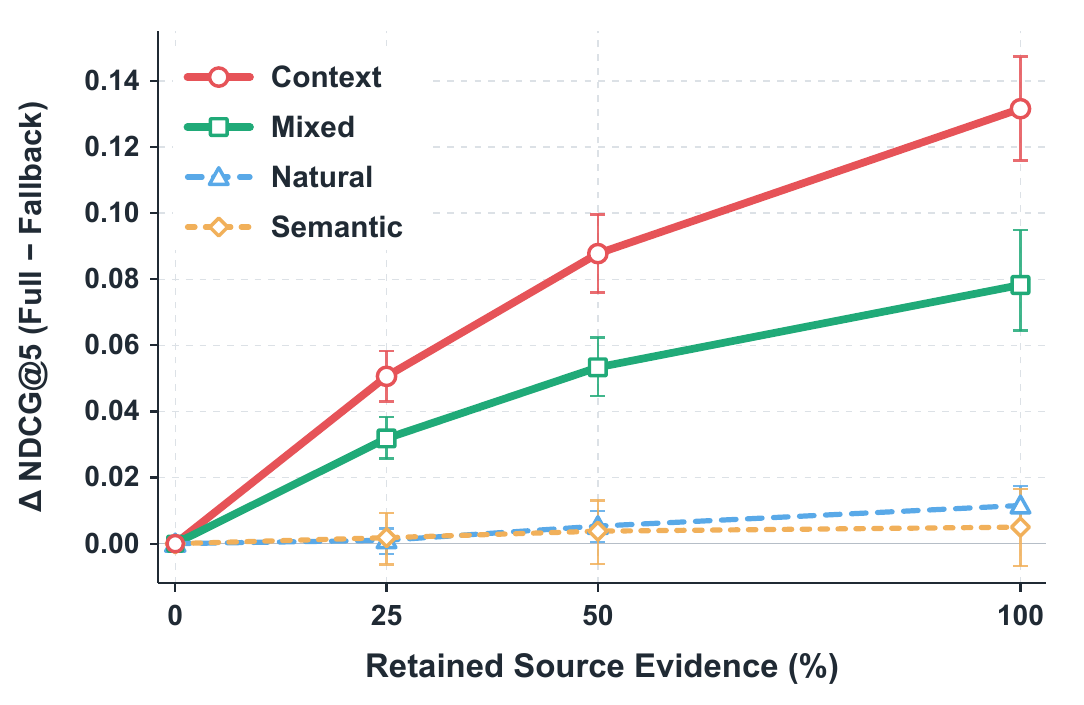}
    \caption{
\textbf{Cross-domain} NDCG@5 gains from contextual validation across
source-association retention levels.  Error bars indicate paired 95\%
cluster-bootstrap intervals.
}
    \label{fig:evidence_scaling}
\end{figure}

\subsection{Scalability and Limits of Transfer Validation (RQ4)}
\label{sec:rq4_robustness}

RQ4 uses the primary experiment's Full checkpoints and shared calibration.

\paragraph{Candidate-library scaling.}
TACT remains ahead of Semantic Cosine as the candidate library grows.
Table~\ref{tab:candidate_scaling} expands the ToolBench library from
the base setting to all 3,442 capabilities.
At full scale, NDCG@5 is 0.401 versus 0.382 on Natural and
0.260 versus 0.056 on Context.
Both methods lose accuracy as more candidates compete for the top ranks,
while TACT's advantage remains larger on Semantic, Context, and Mixed.
Cross-dataset results and candidate-pool definitions appear in
Appendix~\ref{app:rq4_robustness}.

\paragraph{Source-evidence availability.}
The contextual contribution grows with retained source evidence,
most clearly on Context and Mixed.
Figure~\ref{fig:evidence_scaling} varies source associations while
holding model parameters, outer splits, candidate pools, and calibration
fixed. Each checkpoint is shared across the four candidate views.
The comparator, \emph{Fixed-Checkpoint Functional  Fallback}, removes the contextual
contribution from the same checkpoint, allowing its ranking contribution
to be measured at each evidence level.
Full and fallback coincide at zero evidence.
Further comparator details appear in Appendix~\ref{app:rq4_robustness}.

\section{Conclusion}
We presented \methodname{} for zero-shot capability transfer,
refining semantic relevance through Endpoint-Grounded Functional
Validation and Evidence-Bounded Contextual Validation.
Across cross-domain and cross-dataset evaluations, \methodname{}
improves aggregate capability ranking, with larger gains on candidate
sets containing functional or contextual ambiguity.
These gains also translate into higher task completion rates across
four frozen agent backbones under cached execution.
The results support combining semantic relevance with functional
compatibility and contextual applicability for capability selection.

\section*{Acknowledgement}
The work described in this paper was partially supported by InnoHK initiative,
The Government of the HKSAR, and Laboratory for AI-Powered Financial Technologies.

\section*{Limitations}
Our evaluation uses fixed capability libraries and held-out domain or dataset
splits. The reported ranking gains therefore do not establish performance
for newly added capabilities, changing endpoint schemas, or live services.
The closed-world relevance annotations treat unjudged candidates as
non-relevant in expanded pools, so the scaling results measure recovery of
known positives rather than complete compatibility assessment. The cached
execution study uses a fixed set of solvable queries and frozen agent
backbones; its completion rates may not transfer to unrestricted tasks,
live API failures, or other agents. Contextual applicability also depends
on source association evidence and falls back
through the gated functional path when that evidence is unavailable.

\section*{AI Disclosure}
AI-assisted tools were used during manuscript preparation for
language editing, proofreading, and improving readability. The authors
independently developed the research methodology, conducted the experiments,
interpreted the results, and verified all technical content. The authors
remain fully responsible for the accuracy, originality, and integrity of
this work.

\bibliography{reference}

\clearpage
\appendix

\begingroup
\raggedbottom
\renewcommand{\topfraction}{0.95}
\renewcommand{\dbltopfraction}{0.95}
\renewcommand{\textfraction}{0.04}
\renewcommand{\floatpagefraction}{0.72}
\renewcommand{\dblfloatpagefraction}{0.72}
\setcounter{topnumber}{4}
\setcounter{dbltopnumber}{4}
\setcounter{totalnumber}{6}
\setlength{\floatsep}{10pt plus 2pt minus 2pt}
\setlength{\dblfloatsep}{10pt plus 2pt minus 2pt}
\setlength{\textfloatsep}{12pt plus 2pt minus 2pt}
\setlength{\dbltextfloatsep}{12pt plus 2pt minus 2pt}
\makeatletter
\setlength{\@fptop}{0pt}
\setlength{\@fpsep}{12pt}
\setlength{\@fpbot}{0pt plus 1fil}
\setlength{\@dblfptop}{0pt}
\setlength{\@dblfpsep}{12pt}
\setlength{\@dblfpbot}{0pt plus 1fil}
\makeatother

\section{ToolBench Motivating Diagnostic Details}
\label{app:motivation_details}

This appendix specifies the two diagnostics in Figure~\ref{fig:source_diagnostics} of the main paper:
functional ambiguity under frozen semantic retrieval and contextual
instability under a minimal change to task conditions. These diagnostics
characterize the transfer-validity gaps motivating \methodname{}; they are
not evaluations of the full ranking model.

\subsection{Diagnostic Scope and Isolation}
\label{app:motivation_protocol}

The functional diagnostic uses frozen semantic scores and public capability
and endpoint information. The contextual diagnostic uses outcome-blind
minimal pairs followed by blinded review and adjudication. Diagnostic
annotations and outcomes are not used to train, calibrate, select, or tune
\methodname{}. These diagnostics answer different questions from the candidate audits
in Appendix~\ref{app:candidate_audits}; their denominators must not be
combined.

\subsection{Functional Ambiguity in Semantic Retrieval}
\label{app:motivation_semantic}

For task $t$, let $\mathcal{C}^{20}_t$ be its frozen semantic Top-20
candidate set. Within this set, $\mathcal{V}_t$ and $\mathcal{I}_t$ denote
candidates with clear functionally valid and invalid labels, respectively.
A task is eligible only if both sets are nonempty. Unresolved candidates
do not establish eligibility. Functional validity concerns whether the
callable interface supports the required operation and inputs.

For each eligible task, define the smallest semantic separation between
functionally discordant candidates:
\begin{equation}
\begin{aligned}
d_t=\min_{\substack{c^+\in\mathcal{V}_t\\c^-\in\mathcal{I}_t}}
\bigl|s_{\mathrm{sem}}(t,c^+)-s_{\mathrm{sem}}(t,c^-)\bigr|.
\end{aligned}
\label{eq:functional_diagnostic}
\end{equation}
Here $s_{\mathrm{sem}}$ is the frozen semantic score, before the learned
functional or contextual corrections. If $\mathcal{T}_{\mathrm{elig}}$
is the eligible task set, the ambiguity rate at threshold $\epsilon$ is
\begin{equation}
R_{\mathrm{amb}}(\epsilon)=
\frac{1}{|\mathcal{T}_{\mathrm{elig}}|}
\sum_{t\in\mathcal{T}_{\mathrm{elig}}}\mathbb{I}[d_t<\epsilon].
\label{eq:app_ambiguity_rate}
\end{equation}
The denominator is 29,178 eligible tasks. Table~\ref{tab:app_diagnostics}
reproduces the thresholds and rates in the main-text diagnostic.
These are task-level ambiguity rates, not ranking accuracies or
domain-macro averages.

\subsection{Contextual Instability under Minimal Changes}
\label{app:motivation_context}

Each minimal pair $((t^{(0)},c),(t^{(1)},c))$ preserves the capability
and callable function while changing one observed task-context condition.
The pair is constructed without conditioning on whether applicability
will flip. This avoids defining a diagnostic population consisting only
of successful counterfactual flips.

Two automated judges assessed all 500 pairs independently: 324
received clear agreement and 176 underwent follow-up review and
author adjudication. Judgments used task texts, capability descriptions,
endpoint schemas, and context values. TACT scores and expected flip
outcomes were excluded from the judgment evidence. This protocol uses
automated assessment with author adjudication, rather than two
independent human annotations.

Let $\mathcal{P}_{\mathrm{valid}}$ contain pairs whose base task is
supported and whose applicability is determinate on both sides. With
$y_j^{(0)},y_j^{(1)}\in\{0,1\}$ denoting the final applicability labels,
the conditional flip rate is
\begin{equation}
R_{\mathrm{flip}}=
\frac{\sum_{j\in\mathcal{P}_{\mathrm{valid}}}
\mathbb{I}[y_j^{(0)}\neq y_j^{(1)}]}
{|\mathcal{P}_{\mathrm{valid}}|}.
\label{eq:contextual_diagnostic}
\end{equation}
Of the 500 constructed pairs, 408 meet this criterion: 62 flip and 346
remain stable, yielding $62/408=15.20\%$ and $346/408=84.80\%$,
respectively. The other 54 pairs have unsupported base tasks and 38
remain indeterminate; neither category is counted as stable.

The reported 95\% cluster-bootstrap interval for the flip rate is
11.57--19.09\%. It uses 20,000 replicates with seed 20261012, resampling
333 clusters defined by capability name and changed context type.
Table~\ref{tab:app_context_types} disaggregates the included pairs by
the changed context condition.

\begin{table}[tbp]
\centering\small
\begin{tabular}{@{}lr@{}}
\toprule
Diagnostic statistic & Value \\
\midrule
Eligible functional-diagnostic tasks & 29,178 \\
$d_t<0.005$ & 31.41\% \\
$d_t<0.01$ & 44.25\% \\
$d_t<0.02$ & 57.14\% \\
$d_t<0.05$ & 73.62\% \\
\midrule
Constructed contextual pairs & 500 \\
Supported-base, determinate pairs & 408 \\
Stable / flipped pairs & 346 / 62 \\
Applicability flip rate & 15.20\% \\
Cluster-bootstrap 95\% CI & 11.57--19.09\% \\
\bottomrule
\end{tabular}
\caption{Functional ambiguity and contextual instability
corresponding to Figure~\ref{fig:source_diagnostics} and Section~\ref{sec:problem_formulation} of the main paper.}
\label{tab:app_diagnostics}
\end{table}

\begin{table}[tbp]
\centering\small
\setlength{\tabcolsep}{4pt}
\begin{tabular}{@{}lrrr@{}}
\toprule
Changed condition & Included & Flips & Rate (\%) \\
\midrule
Access policy & 64 & 6 & 9.38 \\
Execution environment & 68 & 5 & 7.35 \\
Geographic scope & 62 & 8 & 12.90 \\
Locale & 71 & 9 & 12.68 \\
Platform & 73 & 23 & 31.51 \\
Temporal scope & 70 & 11 & 15.71 \\
\midrule
Total & 408 & 62 & 15.20 \\
\bottomrule
\end{tabular}
\caption{Contextual instability by changed condition. Each
denominator contains only supported-base, determinate pairs. These
descriptive rates are not ranking metrics or cross-domain test results.}
\label{tab:app_context_types}
\end{table}

\subsection{Interpretation and Scope}
\label{app:motivation_role}

Functional ambiguity demonstrates that semantic near-ties can conceal
different endpoint-level validity. Contextual instability demonstrates
that applicability can change even when the capability and callable
function remain fixed. Together, they motivate requirement--endpoint
matching and selective contextual adjustment. Neither rate estimates
the accuracy of an endpoint probe or the benefit of a learned correction.

\section{Task Construction and Endpoint Preprocessing}
\label{app:construction_details}

\subsection{Negative Roles and Candidate Manifests}

Each source negative candidate is assigned one of three roles: ordinary,
semantic, or contextual. Positive candidates retain their positive labels. Ordinary negatives are generally irrelevant candidates. Semantic
negatives are topically related but violate an operation, required-input,
schema, or functional-boundary requirement. Contextual negatives remain
functionally plausible but are unsupported under the relevant source context.

Contextual negatives are excluded from functional-relation training. Ordinary
and semantic negatives are used as controls for contextual adjustment.
In primary ranking experiments, Natural, Semantic, Context, and Mixed
modify only the candidate set $\mathcal{C}_t^{(v)}$; all views share the
same scorer, parameters, gates, and calibration.

\subsection{Task-Requirement Extraction}

A deterministic task-only extractor splits a task into action clauses, attaches
local constraints, merges near-duplicates, and retains at most $N_{\max}=4$
requirements. It reads task text only and never accesses candidate schemas,
source labels, gold endpoints, or target annotations. The same requirement set
is reused for every candidate.

For requirement $n_i$, frozen query vectors are built for operation,
required-argument, and argument-type evidence:
\begin{equation}
\begin{aligned}
q_i^r
&= E_0(\pi_r\oplus n_i),\\
r
&\in\{\mathrm{op},\mathrm{arg},\mathrm{type}\}.
\end{aligned}
\label{eq:task_need_views}
\end{equation}
These are the endpoint-evidence views used in
Section~\ref{sec:functional} of the main paper.

\subsection{Endpoint Preprocessing}

Endpoint rows are deduplicated using the capability identifier, operation name,
description, HTTP method, required argument names, and required argument types.
For each candidate we retain a shared Top-$K$ endpoint set for all functional
evidence channels, with $K=4$ in the reported configuration. This is one shared set per
task--capability pair, rather than a separate Top-$K$ set for each
requirement. All requirements and evidence channels use this same set. Availability
masks $m_j^a$ and $m_j^y$ indicate whether required-argument and argument-type
schema fields are present. Missing fields provide no evidence and are not
interpreted as mismatches.

\subsection{Nested Source-Only Selection}

For each outer target, complete source contexts are held out internally for
checkpoint and strength selection, as detailed in
Appendix~\ref{app:optimization_details}. When a context is used for source
validation, all of its observations are excluded from capability profiles,
support statistics, and task-conditioned source mixtures. After selection, the
chosen configuration is retrained on the complete outer-source set and applied
once to the unseen target.

\section{Relational Compatibility Objective}
\label{app:relational_objective}

The frozen encoder $E_0$ gives normalized task and capability vectors
$e_t,e_c$, and the source-trained relation encoders give normalized states
$z_t,z_c$. The semantic and relational logits are
\begin{equation}
\begin{aligned}
a(t,c)
&=
\frac{\langle e_t,e_c\rangle-\mu_a}{\tau_a},\\
b(t,c)
&=
\frac{\langle z_t,z_c\rangle-\mu_r}{\tau_r}.
\end{aligned}
\label{eq:anchor_calibration_details}
\end{equation}
The frozen anchor and bounded relational refinement are
\begin{equation}
\begin{aligned}
A(t,c)
&=\log\sigma(a(t,c)),\\
L_{\mathrm{rel}}(t,c)
&=a(t,c)+\alpha s_r\\
&\quad\times\tanh\!\left(
\frac{b(t,c)-a(t,c)}{s_r}
\right).
\end{aligned}
\label{eq:app_relational_refinement}
\end{equation}
All calibration temperatures are positive and fitted or selected
using source data only. This reproduces Equations~\eqref{eq:semantic_anchor}--\eqref{eq:relational_compatibility} of the main paper.
Setting $\alpha=0$ recovers the semantic logit; applying $\log\sigma$
then recovers the anchor score.

For a positive capability $c^+$ and a negative capability $c^-$, the pairwise
term is
\begin{equation}
\begin{aligned}
\ell_{\mathrm{rel}}
={}&\operatorname{softplus}\!\Bigl(
m_{\mathrm{rel}}-L_{\mathrm{rel}}(t,c^+)\\
&\hspace{5.2em}+L_{\mathrm{rel}}(t,c^-)
\Bigr).
\end{aligned}
\label{eq:rel_pairwise_loss}
\end{equation}
Semantic false friends receive an additional hardness factor based on frozen
semantic similarity,
\begin{equation}
w_{\mathrm{ff}}
=
\sigma\!\left(
\frac{
\langle e_t,e_{c^-}\rangle
-
\langle e_t,e_{c^+}\rangle
}{T_{\mathrm{ff}}}
\right).
\label{eq:false_friend_weight}
\end{equation}
The complete relation objective combines source-only pairwise supervision with
the false-friend term and the regularizers used by the relation encoder:
\begin{equation}
\mathcal{L}_{\mathrm{rel}}
=
\mathcal{L}_{\mathrm{pair}}
+
\lambda_{\mathrm{ff}}\mathcal{L}_{\mathrm{ff}}
+
\mathcal{R}_{\mathrm{rel}}.
\label{eq:complete_relational_objective}
\end{equation}
Here $\mathcal{R}_{\mathrm{rel}}$ collects the pointwise,
representation-margin, cross-context consistency, and anchor-preservation
terms. Contextual counterfactuals are not used as negatives for this stage.

\section{Functional Validation Details}
\label{app:functional_details}

\subsection{Functional Calibration Reference}
\label{app:single_witness_details}

Requirement--endpoint matching produces endpoint-set support
$C_{\mathrm{ws}}(t,c)$, calibrated as
\begin{equation}
z_{\mathrm{ws}}(t,c)=
\frac{C_{\mathrm{ws}}(t,c)-\mu_{\mathrm{ws}}}{\tau_{\mathrm{ws}}},
\qquad \tau_{\mathrm{ws}}>0.
\label{eq:witness_functional_logit}
\end{equation}
The center and temperature use source data only. The calibrated
endpoint-set score is compared with $L_{\mathrm{rel}}$ to form the
functional residual.

\subsection{Endpoint-Set Matching, Reliability, and Supervision}
\label{app:witness_details}

\paragraph{Endpoint-local evidence.}
Operation, required-argument, and argument-type evidence are composed
within the same endpoint before assignment:
\begin{equation}
v_{ij}=\frac{o_{ij}+\lambda_a m_j^a a_{ij}
+\lambda_y m_j^y y_{ij}}
{1+\lambda_a m_j^a+\lambda_y m_j^y}.
\label{eq:app_joint_endpoint_evidence}
\end{equation}
Missing schema fields are masked out rather than counted as mismatches.
The weighted combination measures compatibility using the fields
available for one endpoint. It keeps operation and argument evidence
from different interfaces separate.

\paragraph{Partial injective assignment.}
Let $M_t\leq N_{\max}$ be the number of extracted requirements and
$K_c\leq K$ the number of retained valid endpoints. The feasible set is
\begin{equation}
\begin{aligned}
\mathcal{X}(t,c)=\{X\in\{0,1\}^{M_t\times K_c}:\quad\\
\sum_j X_{ij}\leq1\ (\forall i),\quad
\sum_i X_{ij}\leq1\ (\forall j)\}.
\end{aligned}
\label{eq:injective_constraints}
\end{equation}
For $M_t>0$, write $|X|=\sum_{i,j}X_{ij}$ and define
\begin{equation}
J(X;t,c)=\frac{\sum_{i,j}X_{ij}v_{ij}
+\nu(M_t-|X|)}{M_t},
\label{eq:witness_assignment_objective}
\end{equation}
\begin{equation}
C_{\mathrm{ws}}(t,c)=\max_{X\in\mathcal{X}(t,c)}J(X;t,c).
\label{eq:witness_coverage}
\end{equation}
Unmatched requirements use the shared null score $\nu$. The number of
feasible assignments is
\begin{equation}
|\mathcal{X}|=\sum_{m=0}^{\min(M_t,K_c)}
\binom{M_t}{m}\binom{K_c}{m}m!.
\label{eq:assignment_count}
\end{equation}
With $N_{\max}=K=4$, the largest assignment bank contains 209 assignments,
including the all-null assignment. Exact enumeration is used during
training and inference. Padded endpoints are not eligible real matches.
If no requirements or endpoints are available, the functional residual
is disabled and no division by an empty requirement count is performed.

\paragraph{Reliability and bounded refinement.}
Let $X^\star$ be a maximizing assignment, $n_X=\sum_{ij}X^\star_{ij}$,
and $m_j^a,m_j^y$ the required-argument and argument-type availability
masks. Structural reliability is defined as
\begin{equation}
\begin{aligned}
q_{\mathrm{schema}}
&=\frac{\sum_{ij}X^\star_{ij}(m_j^a+m_j^y)}
        {2\max(n_X,1)},\\
\rho_{\mathrm{ws}}
&=\mathbb{I}[M_t>0,\ K_c>0]\\
&\quad\times\frac{1+q_{\mathrm{schema}}}{2}.
\end{aligned}
\label{eq:app_structural_reliability}
\end{equation}
Here $M_t$ and $K_c$ count valid requirements and endpoints, excluding
padding. Missing requirements or endpoints disable the residual. If
both are available but the maximizing assignment is all-null,
$\rho_{\mathrm{ws}}=1/2$; all-null matching is not equivalent to
unavailable evidence.

Define $g_{\mathrm{ws}}=z_{\mathrm{ws}}-L_{\mathrm{rel}}$. The functional residual
uses an all-null guard:
\begin{equation}
\widetilde g_{\mathrm{ws}}=
\begin{cases}
g_{\mathrm{ws}}, & n_X>0,\\
\min(g_{\mathrm{ws}},0), & n_X=0.
\end{cases}
\label{eq:app_all_null_guard}
\end{equation}
Its bounded refinement is
\begin{equation}
\begin{aligned}
\Delta_I(t,c)&=\rho_{\mathrm{ws}}(t,c)\widetilde g_{\mathrm{ws}}(t,c),\\
\Gamma_I(t,c)&=\gamma_I s_I
\tanh\!\left(\frac{\Delta_I(t,c)}{s_I}\right),
\end{aligned}
\label{eq:witness_set_correction}
\end{equation}
where $s_I>0$ and $\gamma_I\geq0$. This is the functional residual
and correction in Equations~\eqref{eq:functional_residual}
and~\eqref{eq:functional_correction} of the main paper. The all-null boundary permits
downward correction but prevents an upward correction based entirely
on unmatched requirements. The resulting functional path is
\begin{equation}
\begin{aligned}
L_I(t,c)&=L_{\mathrm{rel}}(t,c)+\Gamma_I(t,c),\\
I(t,c)&=\log\sigma(L_I(t,c)).
\end{aligned}
\label{eq:app_functional_logit}
\end{equation}
Thus $|\Gamma_I|\leq\gamma_I s_I$, and either $\gamma_I=0$ or
$\rho_{\mathrm{ws}}=0$ recovers the preceding relation logit exactly.

\paragraph{Functional-set auxiliary objective.}
For a positive $c^+$ and a negative $c^-$, define
\begin{equation}
\begin{aligned}
\ell_{\mathrm{full}}={}&\operatorname{softplus}\bigl(
m_I-L_I(t,c^+)\\
&\hspace{6em}+L_I(t,c^-)\bigr),
\end{aligned}
\label{eq:witness_full_pair}
\end{equation}
\begin{equation}
\begin{aligned}
\ell_{\mathrm{set}}={}&\operatorname{softplus}\bigl(
0.10-C_{\mathrm{ws}}(t,c^+)\\
&\hspace{6em}+C_{\mathrm{ws}}(t,c^-)\bigr).
\end{aligned}
\label{eq:witness_set_pair}
\end{equation}
Semantic false friends use $0.75\ell_{\mathrm{full}}+0.25\ell_{\mathrm{set}}$;
ordinary negatives use $\ell_{\mathrm{full}}$. The auxiliary objective is
\begin{equation}
\begin{aligned}
\mathcal{L}_{\mathrm{ws}}
={}&\mathcal{L}_{\mathrm{ws,sem}}
+0.30\mathcal{L}_{\mathrm{ws,ord}}\\
&+0.20\mathcal{L}_{\mathrm{ws,pos}}
+0.05\mathcal{L}_{\mathrm{ws,cons}}\\
&+0.002\mathcal{L}_{\mathrm{ws,eng}}.
\end{aligned}
\label{eq:witness_complete_objective}
\end{equation}
The remaining terms stabilize positives, encourage consistency across
source controls, and limit residual magnitude. Only endpoint-set
parameters are updated in this stage. Contextual counterfactuals are
excluded from functional-relation training.

\section{Contextual Validation Details}
\label{app:context_details}

\subsection{Source Profiles and Task Context}

Let $\mu_k$ be the normalized task centroid of source context $k$, and let
$n_{c,k}$ count distinct positive uses of capability $c$ in that context. The
capability profile and task-conditioned source mixture are
\begin{equation}
\begin{aligned}
p_c
&=
\operatorname{norm}\!\left(
\frac{\sum_k\log(1+n_{c,k})\mu_k}
{\sum_k\log(1+n_{c,k})+\epsilon}
\right),\\
\alpha_k(t)
&=
\frac{\exp(\langle e_t,\mu_k\rangle/\tau_{\mathrm{ctx}})}
{\sum_j\exp(\langle e_t,\mu_j\rangle/\tau_{\mathrm{ctx}})},\\
h_t
&=
\operatorname{norm}\!\left(\sum_k\alpha_k(t)\mu_k\right).
\end{aligned}
\label{eq:source_profile_and_context}
\end{equation}
When a source context is held out for inner validation, it is excluded
from all centroids, profiles, and routing statistics.

\paragraph{Cross-dataset profile transfer.}
Source capability identity is the pair $(s,c_s)$, where $s$ denotes a
source dataset and $c_s$ its local capability ID. For each pair, positive
records in the source dataset's Natural view identify the task domains
in which that capability is used. These records are aggregated into a
distribution over contextual prototypes. The profile bank contains
only source datasets; the held-out target dataset contributes no positive
records to it. Shared capability IDs and normalized-name matching are
not required.

To score a target capability $c$, its embedding retrieves the most
similar source capability profiles within each source dataset.
The retrieved profiles are combined with weights reflecting embedding
similarity and source-support strength. The resulting distribution
provides the transferred context profile and its source-support evidence.
If no source candidate exists, or the highest similarity is below $0.25$,
the transferred profile is zero. Otherwise, a capability with a new
target-side ID can inherit evidence from semantically similar source
capabilities. Such evidence remains source-derived; embedding similarity
does not create target-positive observations.

\subsection{Cross-Fitted Evidence Confidence}

Support strength and source-context coverage are
\begin{equation}
\begin{aligned}
q_{\mathrm{str}}(c)
&=
\min\!\left(
1,
\frac{\log(1+\sum_k n_{c,k})}{\log N_0}
\right),\\
q_{\mathrm{cov}}(c)
&=
\min\!\left(
1,
\frac{|\{k:n_{c,k}>0\}|}{K_0}
\right),\\
q_{\mathrm{sup}}(c)
&=
\sqrt{q_{\mathrm{str}}(c)q_{\mathrm{cov}}(c)}.
\end{aligned}
\label{eq:source_support_confidence}
\end{equation}
Routing confidence is
\begin{equation}
q_{\mathrm{route}}(t)
=
1-
\frac{H(\alpha(t))}{\log K_t},
\label{eq:routing_confidence}
\end{equation}
where $K_t>1$ is the number of available source contexts. The
expression is evaluated only when that condition holds; absent source
evidence disables the contextual correction. In the main-text notation,
\begin{equation}
q_{\mathrm{conf}}(t,c)
=
\sqrt{
q_{\mathrm{sup}}(c)\,q_{\mathrm{route}}(t)
}.
\label{eq:context_confidence}
\end{equation}
When neither an observed source-positive profile nor an admissible
transferred source profile is available, set $p_c=0$ and
$q_{\mathrm{sup}}(c)=0$, disabling contextual correction. In
cross-dataset transfer, this condition is evaluated after semantic
profile retrieval, not by testing whether the target capability ID
appears in a source dataset. Coordinate alignment alone supplies no
positive source-use evidence.

\subsection{Centered Task-Conditioned Dynamics}

Let $\bar\mu_{\mathrm{src}}$ be the normalized mean source centroid. The
centered context code is
\begin{equation}
v_{\mathrm{ctx}}(t)
=
E_\Delta(h_t)-E_\Delta(\bar\mu_{\mathrm{src}}).
\label{eq:centered_context_code}
\end{equation}
A low-rank operator gives
\begin{equation}
\delta_0(t,c)
=
U\!\left[
G(v_{\mathrm{ctx}}(t))\odot Vz_c
\right].
\label{eq:context_low_rank_operator}
\end{equation}
We remove the component parallel to $z_c$,
\begin{equation}
\delta_\perp(t,c)
=
\delta_0(t,c)
-
\langle\delta_0(t,c),z_c\rangle z_c.
\label{eq:tangent_projection}
\end{equation}
The source profile contributes a second tangent direction,
\begin{equation}
\begin{aligned}
u_c
&=
\tanh\!\left(A_{\mathrm{use}}(p_c)\right),\\
u_{c,\perp}
&=
u_c-\langle u_c,z_c\rangle z_c.
\end{aligned}
\label{eq:usage_profile_adapter}
\end{equation}
Define the task-observable context activity
\begin{equation}
q_{\mathrm{shift}}(t,c)
\equiv
a_{\mathrm{ctx}}(t)
=
\frac{\|h_t-\bar\mu_{\mathrm{src}}\|_2}
{\|h_t-\bar\mu_{\mathrm{src}}\|_2+1},
\label{eq:context_activity}
\end{equation}
and
\begin{equation}
\operatorname{Bound}(x)
=
\frac{x}{\|x\|_2+\tau_{\mathrm{dev}}},
\label{eq:bounded_mapping}
\end{equation}
where $\tau_{\mathrm{dev}}>0$ is a smoothing constant that keeps
the mapping well-defined at $x=0$ and ensures
$\|\operatorname{Bound}(x)\|_2<1$.
The contextual deviation and task-conditioned capability state are
\begin{equation}
\begin{aligned}
\delta(t,c)
={}&\delta_\perp(t,c)
+\eta_{\mathrm{use}}q_{\mathrm{shift}}(t,c)\\
&\quad\times\operatorname{Bound}(u_{c,\perp}),\\
z_c^{\mathrm{ctx}}(t)
={}&\operatorname{norm}\!\Bigl(
z_c\\
&\quad+\kappa\operatorname{Bound}(\delta(t,c))
\Bigr).
\end{aligned}
\label{eq:contextual_capability_state}
\end{equation}
The relation change in Equation~\eqref{eq:context_relation_change} of the main paper is
\begin{equation}
\Delta r(t,c)=\langle z_t,z_c^{\mathrm{ctx}}(t)\rangle
-\langle z_t,z_c\rangle.
\label{eq:app_context_relation_change}
\end{equation}
Here $q_{\mathrm{shift}}$ is a task-observable context-activity proxy.
Although written as a task--capability quantity for consistency with the
main text, this parameterization depends on the task alone. It is not a
target-context identifier or a directly observed applicability-flip label.

\subsection{Selective Reliability and Evidence Calibration}

The selective reliability in the main paper is therefore
\begin{equation}
\rho_{\mathrm{ctx}}(t,c)
=
q_{\mathrm{conf}}(t,c)
q_{\mathrm{shift}}(t,c).
\label{eq:app_selective_reliability}
\end{equation}
The evidence gate uses source-support, routing, and relation features:
\begin{equation}
g_{\mathrm{ev}}(t,c)
=
1+s_g
\tanh\!\left(
\operatorname{MLP}_D(x_D)
\right).
\label{eq:evidence_gate}
\end{equation}
For $0\leq s_g\leq1$, the gate lies in $[1-s_g,1+s_g]$. Its final
layer is zero-initialized, so $g_{\mathrm{ev}}=1$ before contextual
training.

The contextual correction is
\begin{equation}
\begin{aligned}
D(t,c)
={}&\gamma_D\rho_{\mathrm{ctx}}(t,c)g_{\mathrm{ev}}(t,c)\\
&\times\tanh\!\left(\frac{\Delta r(t,c)}{T_D}\right),
\end{aligned}
\label{eq:app_contextual_correction}
\end{equation}
with $\gamma_D\geq0$ and $T_D>0$, matching
Equation~\eqref{eq:contextual_correction}.
Since $0\leq\rho_{\mathrm{ctx}}\leq1$, the magnitude is bounded by
$|D(t,c)|\leq\gamma_D(1+s_g)$. The correction vanishes when source
confidence or task-observable context activity is zero. Otherwise, its
sign follows $\Delta r$.

The neutrality penalty in the main paper is exactly
\begin{equation}
\mathcal{L}_{\mathrm{neutral}}
=
\mathbb{E}\!\left[
(1-q_{\mathrm{shift}}(t,c))D(t,c)^2
\right].
\label{eq:app_context_neutrality}
\end{equation}

\subsection{Contextual Training Objective}

For a context-supported positive $c^+$ and a contextual counterfactual $c^-$,
we use
\begin{equation}
\begin{aligned}
\mathcal{L}_{\mathrm{cf}}
={}&\mathbb{E}\!\Bigl[
\operatorname{softplus}\!\Bigl(
m_D-D(t,c^+)\\
&\hspace{7.2em}+D(t,c^-)
\Bigr)
\Bigr].
\end{aligned}
\label{eq:context_counterfactual_loss}
\end{equation}
Positive stability discourages negative corrections on supported positives;
ordinary and semantic negatives act as controls against spurious contextual
corrections. The neutrality objective in Equation~\ref{eq:app_context_neutrality}
is included to discourage changes on context-neutral tasks. Fusion losses
are defined in Appendix~\ref{app:fusion_details} and use source pairs only.

\section{Validated Ranking and Analysis Variants}
\label{app:fusion_details}

The reported scorer uses a shared dual-gate fusion network. Its input contains
the source-derived context feature vector together with compact relation
summaries: the bounded relational residual, semantic cosine, relational cosine,
functional relation score, contextual correction, source-evidence confidence,
and context activity. No candidate-view identifier is provided.

The network produces logits $\ell_I$ and $\ell_D$. For
$b\in\{I,D\}$, the suppress-only gate is
\begin{equation}
h_b
=
\tanh(\ell_b),
\qquad
\omega_b
=
1-
\frac{h_b+|h_b|}{2}.
\label{eq:suppress_only_gate}
\end{equation}
Thus $\omega_I,\omega_D\in[0,1]$. The final score is
\begin{equation}
\begin{aligned}
s(t,c)={}&A(t,c)\\
&+\omega_I(t,c)[I(t,c)-A(t,c)]\\
&+\omega_D(t,c)D(t,c),
\end{aligned}
\label{eq:app_framework_score}
\end{equation}
matching Equation~\eqref{eq:framework_score} of the main paper. ``Suppress-only'' refers to the
amplitudes of these gates. It does not constrain $D$ to be non-positive
and does not imply $s(t,c)\leq I(t,c)$.

If $D=0$, the final score recovers the gated functional path
\begin{equation}
s_{\mathrm{fb}}(t,c)=A(t,c)+\omega_I(t,c)[I(t,c)-A(t,c)].
\label{eq:functional_fallback}
\end{equation}
This is equal to $I$ only when the functional gate is one (or $I=A$).
If both corrections are disabled, the score recovers $A$.

For an algebraic removal from Equation~\eqref{eq:app_framework_score}
that preserves the other gate, the two paths are
\begin{equation}
\begin{aligned}
s_{\mathrm{no\ ctx}}&=A+\omega_I(I-A),\\
s_{\mathrm{no\ func}}&=A+\omega_D D.
\end{aligned}
\label{eq:analysis_variants}
\end{equation}
The raw branch scores $I$ and $A+D$ generally differ from these
gate-preserving diagnostic scores. Neither raw branch scores nor
fixed-checkpoint removals are equivalent to independently trained
branch ablations: retraining can change the retained parameters,
calibration, and selected checkpoint. Primary ranking configurations
are selected using source data and shared across candidate views.
The anchor-only variant uses the same frozen task and capability
representations as Semantic Cosine. Because $A(t,c)$ is a monotonic calibration
of the cosine score, the two induce identical rankings.

\subsection{Fusion Supervision}

For a source pair $(c^+,c^-)$, the fused ranking loss is
\begin{equation}
\begin{aligned}
\mathcal{L}_{F}
={}&\mathbb{E}\!\Bigl[
\operatorname{softplus}\!\Bigl(
m_F-s(t,c^+)\\
&\hspace{7.2em}+s(t,c^-)
\Bigr)
\Bigr].
\end{aligned}
\label{eq:fusion_objective}
\end{equation}
Fusion training also uses a source-only dominance term that
prevents the fused margin from falling below the stronger branch margin, a
routing loss that supervises gate suppression from relative source-side branch
margins, and a gate regularizer that keeps both gates near the unsuppressed
state unless source evidence supports suppression. No target labels or view
identifiers enter these losses.

\section{Source-Only Optimization and Model Selection}
\label{app:optimization_details}

\subsection{Staged Objectives}

The frozen text encoder is never updated. Source-only optimization first
fits the anchor-preserving relational refinement in
Appendix~\ref{app:relational_objective}. Endpoint-set parameters are trained
with the functional objective in Appendix~\ref{app:witness_details}, while
contextual and fusion parameters use the objectives in
Appendices~\ref{app:context_details}--\ref{app:fusion_details}. The preceding
relation path is held fixed during the corresponding refinement stages.
Ordinary negatives and semantic false friends supervise functional validity;
contextual counterfactuals supervise contextual applicability.

\subsection{Nested Source-Only Selection}

For each outer target, complete source contexts are held out internally.
Their observations are excluded from the centroids, capability profiles,
support statistics, and routing evidence used to score those contexts.
Checkpoints, calibration parameters, residual strengths, and fusion
settings are selected on these held-out source contexts. Correction-strength
grids include zero to permit fallback when additional evidence does not
improve source validation.

After selection, the chosen configuration is retrained on the complete
outer-source set and evaluated once on the held-out target. Primary
ranking and independently trained branch ablations share each selected
configuration across Natural, Semantic, Context, and Mixed; the view
identifier is not a scorer input. The source-evidence diagnostic uses the
same Full checkpoints and shared calibration as the primary ranking
experiment, as detailed in Appendix~\ref{app:rq4_robustness}. Numerical settings are
consolidated in Appendix~\ref{app:implementation_details}.

\subsection{Shared Functional and Contextual Modes}

Both transfer protocols use at most four task requirements, one shared
Top-4 endpoint set, within-endpoint operation/argument/type composition,
exact partial injective assignment, a bounded functional residual,
selective contextual reliability, and two bounded fusion gates.
These mechanisms are unchanged across candidate views.

\subsection{Cross-Dataset Label-Free Coordinate Alignment}

Cross-domain and cross-dataset experiments use the same scorer
architecture. Before cross-dataset scoring, task and capability
embeddings are statistically aligned using each dataset's own unlabeled
means and standard deviations. This includes the held-out target's
unlabeled statistics and is therefore transductive preprocessing.
Target relevance labels are excluded from model fitting, calibration,
early stopping, checkpoint selection, and source-profile construction.
The target dataset remains excluded from the source-profile bank.

\section{Evaluation Protocol Statistics and Candidate Audits}
\label{app:evaluation_statistics}
\label{app:protocol_statistics}

This appendix complements the evaluation protocol in
Section~\ref{sec:experiments} of the main paper with the exact outer splits,
domain-level coverage, candidate-manifest statistics,
and independent candidate audits.
We additionally clarify the evaluation boundary and the controls used
to prevent target-side information from influencing model development.

\subsection{Outer Splits and Result Aggregation}
\label{app:outer_splits}

Cross-domain evaluation uses leave-one-domain-out (LODO) transfer
over the 49 ToolBench domains. Each target domain is held out once,
and the other 48 domains provide source data. Cross-dataset evaluation
holds out one of ToolBench, TaskBench, MetaTool, and APIBench, using
the other three datasets as sources. Training, calibration, checkpoint
selection, and early stopping use source data only.

For metric $m$, view $v$, and scored target set $\mathcal{U}_v$, write
\begin{equation}
\overline m_v=\frac{1}{|\mathcal{U}_v|}
\sum_{u\in\mathcal{U}_v}\left(\frac{1}{3}\sum_{s=1}^{3}m_{u,v,s}\right).
\label{eq:app_target_macro}
\end{equation}
The three runs use seeds 42, 43, and 44. A domain or dataset receives
one vote regardless of its task count. For cross-domain evaluation,
$\mathcal{U}_v$ contains domains with evaluation tasks and at least five
available candidate capabilities in view $v$. Domain--view cells with
fewer than five capabilities are excluded from the macro average;
their scores are not filled with zero. This eligibility rule is shared
by all methods and seeds. It gives 49, 39, 49, and 44 scored domains for
Natural, Semantic, Context, and Mixed, respectively.
Per-target means and sample standard deviations are reported in
Appendix~\ref{app:full_per_target}.
Model fitting and selection follow
Appendix~\ref{app:optimization_details}; the outer target supplies no
relevance supervision for these operations.

Table~\ref{tab:dataset_protocol_statistics} summarizes the overall
scale of the two protocols.
Tasks refer to the Natural-view evaluation population, while candidate
pairs count all task--capability records in the corresponding frozen
manifests.

\begin{table*}[tp]
\centering
\small
\setlength{\tabcolsep}{5.5pt}
\renewcommand{\arraystretch}{1.06}

\begin{adjustbox}{max width=\textwidth}
\begin{tabular}{llrrrrrrr}
\toprule
Protocol
& Held-out unit
& Units
& Tasks
& Capabilities
& Endpoints
& Positive pairs
& Candidate pairs
& Cand./Task \\
\midrule

Cross-domain
& ToolBench domain
& 49
& 201,774
& 3,442
& 19,096
& 335,168
& 4,370,648
& 21.66 \\

Cross-dataset
& Dataset
& 4
& 254,672
& 4,881
& 20,535
& 419,632
& 4,941,883
& 19.40 \\

\bottomrule
\end{tabular}
\end{adjustbox}

\caption{
Overall statistics of the cross-domain and cross-dataset evaluation
protocols.
Cross-domain evaluation holds out each of 49 ToolBench domains,
whereas cross-dataset evaluation holds out each of
ToolBench, TaskBench, MetaTool, and APIBench once.
Tasks denote Natural-view evaluation tasks, and capabilities and
endpoints are counted over the corresponding frozen manifests.
}
\label{tab:dataset_protocol_statistics}
\end{table*}

\subsection{Domain Coverage and Evaluation Eligibility}
\label{app:domain_coverage}

The domain roster follows the 49 ToolBench domain names listed in
Tables~\ref{tab:per_domain_natural}--\ref{tab:per_domain_mixed}.
The Natural and Context results contain a score for every domain.
Semantic has records in 42 domains but scores in 39; Mixed has
records in 46 domains but scores in 44. The excluded nonempty cells
contain fewer than five available candidate capabilities: Jobs,
Logistics, and Monitoring in Semantic, and Energy and Storage in Mixed.
The detailed tables retain their task counts and mark ranking scores
with a dash. Here $N$ counts tasks, not capabilities. Candidate-manifest
statistics cover all constructed records, whereas ranking macro averages
include only eligible domains.

\subsection{Candidate-Manifest Statistics}
\label{app:candidate_statistics}

The four candidate views are constructed to probe different ranking
failures.
Natural preserves the original candidate distribution;
Semantic introduces functionally incompatible but semantically related
candidates;
Context introduces candidates whose apparent support does not transfer
safely to the target context;
and Mixed combines the Semantic and Context hard-negative types with ordinary
candidates.

Eligibility is determined separately for each view.
A task is retained only when the required view-specific hard candidate can
be constructed under the frozen rules.
The resulting task populations may therefore differ across views.
Within every dataset--view or group--view manifest, however, all methods
are evaluated on exactly the same tasks and candidate sets.

\paragraph{Cross-dataset candidate composition.}
Table~\ref{tab:per_dataset_view_statistics} reports the scale and
candidate-role composition of each dataset--view combination.
The ``Other'' column contains the remaining ordinary or unclassified
candidate records after positives, Semantic candidates, and Context
candidates are counted.

\begin{table*}[tp]
\centering
\small
\setlength{\tabcolsep}{2.7pt}
\renewcommand{\arraystretch}{1.03}

\begin{tabular}{llrrrrrrrr}
\toprule
Dataset
& View
& Tasks
& Caps.
& Pairs
& Pos. (\%)
& Sem. (\%)
& Ctx. (\%)
& Other (\%)
& Cand./Task \\
\midrule

\multirow{4}{*}{ToolBench}
& Natural
& 201,774 & 3,442 & 4,370,648
& 7.7 & 0.0 & 0.0 & 92.3 & 21.66 \\
& Semantic
& 50,254 & 3,442 & 1,084,884
& 7.4 & 30.4 & 0.0 & 62.3 & 21.59 \\
& Context
& 75,709 & 3,442 & 1,637,800
& 7.5 & 0.0 & 51.4 & 41.1 & 21.63 \\
& Mixed
& 62,266 & 3,442 & 1,345,846
& 7.5 & 22.0 & 24.7 & 45.9 & 21.61 \\
\midrule

\multirow{4}{*}{TaskBench}
& Natural
& 17,136 & 103 & 205,535
& 23.7 & 0.0 & 0.0 & 76.3 & 11.99 \\
& Semantic
& 17,136 & 103 & 220,048
& 22.1 & 77.9 & 0.0 & 0.0 & 12.84 \\
& Context
& 17,136 & 103 & 220,048
& 22.1 & 0.0 & 77.9 & 0.0 & 12.84 \\
& Mixed
& 17,136 & 103 & 220,048
& 22.1 & 31.1 & 31.1 & 15.6 & 12.84 \\
\midrule

\multirow{4}{*}{MetaTool}
& Natural
& 20,544 & 199 & 235,440
& 8.7 & 0.0 & 0.0 & 91.3 & 11.46 \\
& Semantic
& 20,544 & 199 & 202,633
& 10.1 & 89.9 & 0.0 & 0.0 & 9.86 \\
& Context
& 10,807 & 199 & 118,889
& 9.1 & 0.0 & 90.9 & 0.0 & 11.00 \\
& Mixed
& 20,544 & 199 & 225,178
& 9.1 & 36.1 & 19.2 & 35.5 & 10.96 \\
\midrule

\multirow{4}{*}{APIBench}
& Natural
& 15,218 & 1,137 & 130,260
& 11.7 & 0.0 & 0.0 & 88.3 & 8.56 \\
& Semantic
& 15,218 & 1,135 & 154,665
& 9.8 & 90.2 & 0.0 & 0.0 & 10.16 \\
& Context
& 15,218 & 1,135 & 167,398
& 9.1 & 0.0 & 90.9 & 0.0 & 11.00 \\
& Mixed
& 15,218 & 1,137 & 167,295
& 9.1 & 36.3 & 36.4 & 18.2 & 10.99 \\
\bottomrule
\end{tabular}

\caption{
Per-dataset scale and candidate composition of the four views under
cross-dataset leave-one-dataset-out evaluation.
Caps. denotes the number of distinct capabilities appearing in each
view manifest. Percentages are independently rounded and may not sum
to exactly 100.0.
Pos., Sem., and Ctx. denote observed positives, Semantic candidates,
and Context candidates, respectively.
View-specific eligibility conditions may produce different task
populations, but all methods share the same frozen manifest within each
dataset--view combination.
}
\label{tab:per_dataset_view_statistics}
\end{table*}

\paragraph{Cross-domain candidate composition.}
Table~\ref{tab:cross_domain_view_statistics} aggregates the corresponding
statistics over the 49 ToolBench domains.
Natural and Context retain tasks from all 49 underlying domains,
whereas Semantic and Mixed have candidate records in 42 and 46
domains, respectively. Ranking evaluation further requires at least
five available capabilities in a domain--view cell, giving 39 scored
Semantic domains and 44 scored Mixed domains. All methods use the same
eligible domains within each view.

\begin{table*}[tp]
\centering
\small
\setlength{\tabcolsep}{3.7pt}
\renewcommand{\arraystretch}{1.04}

\begin{tabular}{lrrrrrrrrr}
\toprule
View
& Domains
& Tasks/Pools
& Caps.
& Pairs
& Pos. (\%)
& Sem. (\%)
& Ctx. (\%)
& Other (\%)
& Cand./Task \\
\midrule

Natural
& 49
& 201,774
& 3,442
& 4,370,648
& 7.7
& 0.0
& 0.0
& 92.3
& 21.66 \\

Semantic
& 42
& 50,254
& 3,442
& 1,084,884
& 7.4
& 30.4
& 0.0
& 62.3
& 21.59 \\

Context
& 49
& 75,709
& 3,442
& 1,637,800
& 7.5
& 0.0
& 51.4
& 41.1
& 21.63 \\

Mixed
& 46
& 62,266
& 3,442
& 1,345,846
& 7.5
& 22.0
& 24.7
& 45.9
& 21.61 \\

\bottomrule
\end{tabular}

\caption{
Aggregate scale and candidate composition of the four views under the
ToolBench candidate manifests.
Domains denotes the number of domains containing candidate records;
it is not the scored-domain count used in the ranking macro average.
Caps. is the number of distinct candidate capabilities.
Pos., Sem., and Ctx. denote observed positives, Semantic candidates,
and Context candidates, respectively.
}
\label{tab:cross_domain_view_statistics}
\end{table*}

\subsection{Candidate Audits}
\label{app:candidate_audits}

We assess candidate construction with a double-blind, two-stage audit.
Two AI judges review samples from all four datasets, and two human
reviewers independently audit a ToolBench sample. Assessors do not see
method identities, model scores, evaluation outcomes, or one another's
judgments; human reviewers also do not see the AI decisions.
Acceptance within each stage requires both assessors to approve the
record. For records assessed in both stages, final acceptance requires
passing both the AI and human audits. The tables report the stages
separately over their respective samples. Audit labels characterize
the frozen candidate manifests and are not used for training or scoring.

\paragraph{Cross-dataset AI construction audit.}
For each dataset, two independent AI judges evaluate Semantic and Context
candidates as matched task--candidate pairs and evaluate Mixed at the
complete candidate-pool level.
Both judges apply the same view-specific rubric for semantic relevance,
functional compatibility, and contextual applicability. A pair or pool
passes the AI stage only when both judges confirm that it meets the
criteria of its intended view. A rejection by either judge makes the
record invalid; if neither rejects it but a required judgment is
indeterminate, the record is unresolved.
Table~\ref{tab:per_dataset_candidate_validity} reports this audit across
all four datasets and the three constructed candidate views.

\begin{table*}[tp]
\centering
\small
\setlength{\tabcolsep}{4.5pt}
\renewcommand{\arraystretch}{1.04}

\begin{tabular}{llcrrrrr}
\toprule
Dataset
& View
& Unit
& $N$
& Valid
& Valid (\%)
& Invalid
& Unresolved \\
\midrule

\multirow{3}{*}{ToolBench}
& Semantic & Pair & 160 & 152 & 95.0 & 8  & 0 \\
& Context  & Pair & 160 & 148 & 92.5 & 12 & 0 \\
& Mixed    & Pool & 100 & 92  & 92.0 & 8  & 0 \\
\midrule

\multirow{3}{*}{TaskBench}
& Semantic & Pair & 160 & 154 & 96.3 & 6  & 0 \\
& Context  & Pair & 160 & 150 & 93.8 & 10 & 0 \\
& Mixed    & Pool & 100 & 93  & 93.0 & 7  & 0 \\
\midrule

\multirow{3}{*}{MetaTool}
& Semantic & Pair & 160 & 154 & 96.3 & 6  & 0 \\
& Context  & Pair & 160 & 149 & 93.1 & 11 & 0 \\
& Mixed    & Pool & 100 & 91  & 91.0 & 9  & 0 \\
\midrule

\multirow{3}{*}{APIBench}
& Semantic & Pair & 160 & 158 & 98.8 & 2 & 0 \\
& Context  & Pair & 160 & 154 & 96.3 & 6 & 0 \\
& Mixed    & Pool & 100 & 99  & 99.0 & 1 & 0 \\
\bottomrule
\end{tabular}

\caption{
Per-dataset AI construction audit.
Two independent AI judges apply the same view-specific rubric under
double-blind review. Valid counts require approval from both judges.
Semantic and Context are evaluated at the pair level, whereas Mixed is
evaluated at the candidate-pool level.
}
\label{tab:per_dataset_candidate_validity}
\end{table*}
\begin{table*}[tp]
\centering
\small
\setlength{\tabcolsep}{4.3pt}
\renewcommand{\arraystretch}{1.04}

\begin{tabular}{lrrrrrrrr}
\toprule
View
& Audited records
& Unit
& $N$
& Valid
& Valid (\%)
& Agreement (\%)
& Sem. $\kappa$
& Func. $\kappa$ \\
\midrule

Natural
& 98
& Pair
& 85
& 82
& 96.5
& 92.7
& 0.920
& 0.767 \\

Semantic
& 126
& Pair
& 84
& 78
& 92.9
& 89.1
& 0.750
& 0.844 \\

Context
& 147
& Pair
& 98
& 92
& 93.9
& 90.3
& 0.797
& 0.740 \\

Mixed
& 496
& Pool
& 46
& 43
& 93.5
& 93.8
& 0.687
& 0.747 \\

\bottomrule
\end{tabular}

\caption{
Double-blind human audit under ToolBench cross-domain transfer.
Acceptance requires approval from both human reviewers and passage of
the AI audit under the same view-specific criteria.
For Semantic and Context, $N$ denotes audited hard candidates; for
Mixed, it denotes audited candidate pools.
Agreement and Cohen's $\kappa$ are computed from the two independent
human reviews.
Natural is included as a reference for reviewer agreement rather than as
a candidate-view construction test.
}
\label{tab:cross_domain_candidate_audit}
\end{table*}
\paragraph{Independent ToolBench human audit.}
In the second stage, two human reviewers independently assess sampled
ToolBench records using the same view-specific criteria and double-blind
protocol. A record is accepted only if both reviewers approve it and it
also passes the AI stage. Neither stage can compensate for failure in
the other. Table~\ref{tab:cross_domain_candidate_audit} reports acceptance,
raw human agreement, and Cohen's $\kappa$ for the four views. Agreement
statistics use the independent human judgments before combining stages.
Natural is retained as a reference for annotation consistency, whereas
Semantic, Context, and Mixed evaluate the corresponding construction
criteria.

\subsection{Leakage Controls and Evaluation Boundary}
\label{app:evaluation_boundary}

All candidate manifests are frozen before evaluation and shared by all
compared methods.
Target-side task--capability annotations are used only to define final
evaluation records and compute ranking metrics.
They are never used for model training, calibration, early stopping,
source-context selection, training-negative sampling, expansion ranking,
or hyperparameter selection. Their use in constructing frozen evaluation
labels and candidate manifests is kept separate from model development.

At inference, each method may access the task text, permitted public
capability information, and the source-derived information listed in
Table~\ref{tab:method_input_specifications}. Cross-dataset alignment may
also use the unlabeled target embedding statistics specified there.
The target-context identifier and target-side relevance labels are never
provided as model inputs.
When a source context is held out for internal selection, its
observations are excluded from the source-derived profiles and
statistics used to evaluate that context.

Candidate audits assess benchmark quality only and do not alter the
frozen candidate sets or reported results.
In expanded-library experiments, annotated target positives define the
relevance boundary, while additional unjudged capabilities are treated
as non-relevant.
These experiments therefore measure recovery of known positives as the
library grows, rather than establishing that every unannotated
capability is functionally invalid.

\section{Implementation Details}
\label{app:implementation_details}

This section provides the staged optimization protocol, principal
hyperparameters, baseline adaptations, and computational environment
corresponding to the implementation details summarized in the main text.

\subsection{Optimization Protocol}

The staged objectives, cross-fitting boundary, and nested source-only
selection procedure are specified in
Appendix~\ref{app:optimization_details}. This section collects numerical
settings, baseline adaptations, public-input specifications, and the
computational environment. Discrete hyperparameter sets below are
source-only search spaces, not settings selected using the outer target.

\subsection{Main Configuration}

Table~\ref{tab:main_configuration} summarizes the principal TACT
configuration.
\begin{table}[tbp]
\centering
\small
\setlength{\tabcolsep}{3pt}
\renewcommand{\arraystretch}{1.04}

\begin{tabularx}{\columnwidth}{@{}>{\raggedright\arraybackslash}X>{\raggedright\arraybackslash}X@{}}
\toprule
Configuration & Value \\
\midrule
Frozen encoder
& \texttt{all-MiniLM-L6-v2} \\
Embedding dimension
& 384 \\
Maximum requirements $N_{\max}$
& 4 \\
Endpoint retrieval $K$
& 4 \\
Maximum exact assignments
& 209 \\

Initial null score $\nu$
& 0.15 \\

Functional residual scale $s_I$
& 1.0 \\

Endpoint-set temperature $\tau_{\mathrm{ws}}$
& 0.10 \\

Endpoint-set epochs
& $\{0,1,2,3,4\}$ \\

Functional strength $\gamma_I$
& $\{0,.025,.05,\allowbreak .10,.15,.20\}$ \\

Optimizer
& AdamW \\

Learning rate
& $6\times10^{-4}$ \\

Weight decay
& $10^{-4}$ \\

Gradient clipping
& 1.0 \\

Context temperature
& 0.07 \\

Fusion
& Branch-specific gates \\

Selection
& Source contexts only \\
\bottomrule
\end{tabularx}

\caption{
Main implementation configuration of TACT.
Discrete sets denote source-only search spaces.
All correction grids contain a zero-valued setting to permit exact
fallback to the preceding component.
}
\label{tab:main_configuration}
\end{table}
We use a frozen \texttt{all-MiniLM-L6-v2} encoder and do not update its
parameters during training.
The task-side requirement extractor produces at most four requirements, and endpoint
retrieval retains a shared set of at most four endpoints for each
task--capability pair, reused across all extracted requirements and
evidence channels.
Endpoint-set training epochs and correction strengths are selected
using source contexts.

\subsection{Baseline Adaptations and Input Specifications}
\label{app:baseline_details}

All baselines are adapted to the same task--capability ranking
interface.
Given a task and a frozen candidate set, each method produces a scalar
score for every task--capability pair and ranks candidates in descending
order.
All methods use the same outer splits, candidate manifests, and
evaluation labels.
Target-side relevance labels are reserved exclusively for final
evaluation.

Random assigns pseudo-random candidate scores reproducibly within
each evaluation seed.
Semantic Cosine encodes the same public task and capability text used by
the frozen \methodname{} anchor and ranks candidates by cosine similarity.
It requires no supervised task--capability fitting.

Endpoint Schema Match is a deterministic, training-free structural
baseline.
Within the operation-retrieved endpoint pool, it aggregates
argument-name and argument-type similarities for each endpoint and uses
the maximum endpoint score for ranking.
It receives no relevance labels, context identifiers, or outputs from
\methodname{}.

Logistic Ranker applies pointwise classification to compact
task--capability comparison features, including semantic similarity,
embedding distance, element-wise interactions, and representation
norms.
Pointwise MLP takes the task embedding, capability embedding, their
element-wise product, and their absolute difference as input.

MultiPool pools all outer-source domains or datasets and trains one
ranker on the resulting source data. It has no explicit mechanism for
separating invariant compatibility from context-dependent applicability.
For cross-dataset evaluation, the MultiPool variant further
applies label-free transductive model-signature alignment using unlabeled
target task and capability embedding statistics. It never uses target
relevance labels. \methodname{} also uses unlabeled target embedding
statistics for label-free coordinate alignment, as stated in the main
text; the two alignment procedures should not be conflated.

ZESRec, UniSRec, and AREAD adapt the corresponding transfer-learning
ideas to task--capability ranking.
ZESRec uses separate task and capability towers with source-trained
capability residuals.
UniSRec applies mixture-of-experts adaptors to frozen task and
capability representations.
AREAD combines an invariant score with a source-context-aware
correction.

Re-Invoke follows the original configuration with
\texttt{text-bison@001} for capability-side query generation and task-side
intent extraction, and \texttt{textembedding-gecko@003} for dense
retrieval \citep{chen2024reinvoke}.
ToolDreamer uses GPT-4.1 to generate hypothetical tools and the authors'
Qwen3-8B dense and BM25 retrieval configurations
\citep{sengupta2026tooldreamer}. The generated thought, tool name, and
description are combined with the task query (QTND), and retrieved lists
are combined by reciprocal rank fusion (RRF).
These retrieval configurations use public task and capability text
without target relevance labels or source-side retriever fitting.
\paragraph{PORTS.}
We use the authors' PORTS implementation\footnote{\url{https://github.com/disi-unibo-nlp/ports}}
with its original backbone and encoder configurations
\citep{molfetta2025ports}. The frozen preference models are
Codestral-22B-v0.1, Llama-3-8B, and Llama-3-Groq-8B-Tool-Use;
the retriever encoders are RoBERTa-base and BGE-base.
The adaptation maps task--capability records to the authors' retrieval
interface and fits the retriever with preference supervision from
outer-source records. Model selection also uses source data only.
Results use seeds 42, 43, and 44.

\paragraph{Baseline configurations.}
Table~\ref{tab:baseline_hyperparameters} reports model architectures and
optimization settings. Pointwise MLP, ZESRec, UniSRec, AREAD, PORTS,
and MultiPool use outer-source training data. The table also specifies
the Logistic Ranker control. Generation-enhanced retrievers use the
fixed configurations below.

\begin{table*}[tp]
\centering
\scriptsize
\setlength{\tabcolsep}{4.5pt}
\renewcommand{\arraystretch}{1.08}
\begin{tabularx}{\textwidth}{@{}p{0.17\textwidth}X@{}}
\toprule
\textbf{Method} & \textbf{Configuration} \\
\midrule
Logistic Ranker &
Eight pair features; \texttt{StandardScaler}; balanced pointwise logistic
regression with $C=1$, \texttt{lbfgs}, and 500 maximum iterations. \\

Pointwise MLP &
Input $[e_t,e_c,e_t\odot e_c,|e_t-e_c|]$;
$4D\!\rightarrow\!256\!\rightarrow\!64\!\rightarrow\!1$ with LayerNorm,
ReLU, and dropout $0.10$; weighted BCE with positive weight $\min(n_{-}/n_{+},20)$; AdamW
($10^{-3}$, weight decay $10^{-4}$); batch size 2048; cosine schedule;
clip 1.0; optimized for up to 20 cross-domain or 12 cross-dataset
epochs. \\

MultiPool &
Separate $D\!\rightarrow\!256\!\rightarrow\!128$ task/capability encoders,
dropout $0.05$, raw-cosine plus zero-initialized bounded residual;
pairwise hinge margin $0.03$; AdamW ($10^{-3}$, $10^{-4}$); at most
50,000 pairs, two positives and five negatives per task; batch size
1024/2048 and up to 20/12 training epochs for
cross-domain/cross-dataset. The cross-dataset rows use
label-free transductive \texttt{model\_signature} alignment
based on unlabeled target embedding statistics. \\

ZESRec &
Separate $D\!\rightarrow\!256\!\rightarrow\!128$ towers; normalized dot
product; 128-d zero-initialized capability-ID residual (weight 0.20) and
ID bias (weight 0.05); no domain ID. Weighted BCE and the Pointwise-MLP
optimizer, batch size, schedule, clipping, and training budget. \\

UniSRec &
Separate $D\!\rightarrow\!128$ MoE adaptors with eight parametric-whitening
experts, dropout $0.05$, noisy soft gates, and learned temperature
initialized at 10 and clipped to $[1,30]$; no domain ID. The same weighted
BCE objective and optimization schedule as Pointwise MLP. \\

AREAD &
Invariant and domain towers $D\!\rightarrow\!192\!\rightarrow\!96$;
64-d source-domain embedding, FiLM conditioning, four experts,
domain-correction gate, $\alpha=1$, and centroid inference temperature
$0.10$. It uses the same weighted BCE objective and optimization
schedule as Pointwise MLP. \\

Re-Invoke &
Original configuration; \texttt{text-bison@001} for query generation
and intent extraction; \texttt{textembedding-gecko@003} encoder.
Ten synthetic queries per capability with temperature 0.7; task-intent
extraction with temperature 0.0; no source-side fitting. \\

PORTS &
Authors' implementation; frozen preference models:
Codestral-22B-v0.1, Llama-3-8B, and Llama-3-Groq-8B-Tool-Use;
retriever encoders: RoBERTa-base and BGE-base.
Preference-based retriever adaptation and model selection use
outer-source data; seeds 42, 43, and 44. \\

ToolDreamer &
Original configuration; GPT-4.1 hypothetical-tool generator;
Qwen3-8B dense retriever and BM25; QTND input construction and RRF;
no source-side retriever fitting or target-label selection. \\
\bottomrule
\end{tabularx}
\caption{Configurations of baselines and controls. ZESRec, UniSRec,
and AREAD are task-to-capability adaptations. Re-Invoke, ToolDreamer,
and PORTS retain the authors' model configurations and use the same
task--capability ranking interface.}
\label{tab:baseline_hyperparameters}
\end{table*}

\paragraph{Generation settings and prompts.}
Table~\ref{tab:generation_prompts} summarizes the generation and retrieval
configurations. The generation prompts are given below. In Re-Invoke, \texttt{\{tool\_document\}} is the normalized
public capability text; no separate endpoint or parameter-schema columns
are appended.

\begin{table*}[tp]
\centering
\scriptsize
\setlength{\tabcolsep}{4.5pt}
\renewcommand{\arraystretch}{1.08}
\begin{tabularx}{\textwidth}{@{}p{0.17\textwidth}X@{}}
\toprule
\textbf{Method} & \textbf{Generation and retrieval configuration} \\
\midrule
Re-Invoke &
\texttt{text-bison@001} generates ten synthetic queries per capability
at temperature 0.7 and extracts task intents at temperature 0.0.
\texttt{textembedding-gecko@003} embeds augmented documents and intents;
the expanded document embeddings are averaged. \\
ToolDreamer &
GPT-4.1 generates hypothetical tools. Each tool contributes its thought,
name, and description alongside the original query (QTND).
Retrieval uses the Qwen3-8B dense and BM25 configurations with RRF
over the resulting ranked lists. \\
\bottomrule
\end{tabularx}
\caption{Generation and retrieval settings used by Re-Invoke and ToolDreamer.}
\label{tab:generation_prompts}
\end{table*}

\subparagraph{Re-Invoke tool-side prompt.}
\begingroup\footnotesize
\begin{Verbatim}[breaklines=true,breakanywhere=true]
Suppose you are an assistant and you have access to the following API to answer user's queries.
You are provided with a tool and its available API function including the description and
parameters.
Your task is to generate a possible user query that can be handled by the API.
You must include the input parameters required in the API call. Please be creative and generate
random but specific information for the required parameters.
Now you are given the API documentation below:
{tool_document}
Please generate a user query that you will need to call this tool. Note the generated query
should be complex enough to describe the scenarios that you will need to call the provided API
to address them.
The relevant query is:
\end{Verbatim}
\endgroup
The ten generated queries are serialized as
\texttt{Documentation: \{skill\_text\} Query: \{synthetic\_query\}}.
Their \texttt{textembedding-gecko@003} vectors are averaged to
represent the augmented capability document.
Multiple task intents are fused by the maximum lexicographic
$(\text{reversed rank},\text{similarity})$ tuple.

\subparagraph{Re-Invoke task-side intent prompt.}
\begingroup\footnotesize
\begin{Verbatim}[breaklines=true,breakanywhere=true]
**Instructions**
Suppose you are a query analyzer and your task is to extract the underlying user intents from
the input query. You should preserve all the underlying user request and the extracted user
intents should be easily understood without extra context information.
You should carefully read the given user query to understand its different intents. Then
identify what are the specific intents. Each individual intent should be separated by a newline.
Here are some examples of how you should solve the task.
Example
Query: I'm planning to travel to Paris next weekend to visit my family, could you help me book a
round trip flight ticket? I want to fly in economy class.
Intent:
book a round-trip flight ticket in economy class to Paris next weekend
Query: I'm a potential buyer looking for a condominium in the city of Miami. I am specifically
interested in properties that have a minimum of two bathrooms. It should have walkable distance
to the grocery stores.
Intent:
buy a real estate in Miami with a minimum of two bathrooms and walkable distance to the grocery
stores
Query: I want to learn Spanish by talking to the native speakers at any time. Additionally, can
you recommend some interesting books, preferably fictions, so that I can learn by reading? Also
include the websites that I can buy them.
Intent:
learn Spanish by talking to the native speakers
recommend fictions to learn Spanish by reading
suggest the websites to buy Spanish fictions
**Begin!**
Query: {user_query}
Intent:
\end{Verbatim}
\endgroup

\subparagraph{ToolDreamer prompt.}
\begingroup\footnotesize
\begin{Verbatim}[breaklines=true,breakanywhere=true]
You are a tool-crafting assistant. Your goal is to design a set of hypothetical tools that can
collectively solve a given user query. These tools behave like independent, stateless functions
and have access to all relevant structured data needed to complete their task. No external
internet access is assumed.
High-Level Strategy
1. Break down the query into distinct, logically separable subtasks.
2. For each subtask, propose exactly one tool that performs only that task.
3. Tools must be modular, reusable, and generic where possible (e.g., 'getPopulationData'). If a
subtask clearly requires specialized or scoped logic (e.g., region-specific), it's acceptable to
use a specialized tool, but its name must reflect its limited scope (e.g.,
'weatherAPI_California', 'fetchStockData_TechSector').
4. All tools must be:
- Implementation-agnostic
- Independent
- Named consistently using 'camelCase' or 'snake_case'
5. Every query must result in at least one tool being proposed.
Output Format
For each tool, respond using the following format:
Thought: [Explain why this tool is necessary and what subtask it addresses]
Tool Name: [Concise and consistent tool name]
Tool Description: [Describe what this tool does in one or two sentences, avoiding implementation
details]
Example
User Query:
What is the average temperature in San Francisco?
Assistant Response:
Thought: To answer this query, we need to retrieve historical and/or current temperature data
for a specific location in California.
Tool Name: weatherAPI_California
Tool Description: Retrieves current and historical temperature data for locations in California.
 Additional Tips
- Do not suggest tool pipelines or combine subtasks into a single tool.
- If the query involves data transformation (e.g., averaging, filtering, ranking), propose a
separate tool for that.
- Treat each tool as stateless and composable.
User Query:
{user_query}
Assistant Response:
\end{Verbatim}
\endgroup
Each parsed tool is serialized into the following fields for retrieval:
\begin{Verbatim}[fontsize=\footnotesize,breaklines=true,breakanywhere=true]
Question: {original_query}
Thoughts: {thought}
Tool Name: {name}
Tool Description: {description}
\end{Verbatim}
The retrieval configurations use Qwen3-8B dense embeddings and BM25
scores. Ranked lists from the hypothetical tools are combined with RRF,
and direct-query retrieval supplies candidates outside their union and
the fallback when generation or parsing fails.

\paragraph{Input parity.}
Table~\ref{tab:method_input_specifications} summarizes the information
available to representative comparison methods.
In particular, Endpoint Schema Match is given exactly the same public
endpoint-operation and argument-name/type schema evidence as TACT.
Therefore, the comparison controls for access to public endpoint/schema
evidence rather than advantages from additional schema inputs.
The main-paper w/o Functional ablation removes Functional Validation,
while w/o Context removes Contextual Validation.
TACT further decomposes the task into local requirements, requires supporting
evidence to be composed within the same endpoint, imposes
capacity-constrained requirement–endpoint matching, and calibrates the resulting
evidence using structural reliability and bounded corrections.
TACT additionally uses source-derived capability profiles in its
contextual applicability branch; these profiles are not used by Endpoint
Schema Match.

All trainable baselines---Logistic Ranker, Pointwise MLP, ZESRec,
UniSRec, AREAD, PORTS, and MultiPool---use outer-source training data only.
Cross-dataset MultiPool additionally uses unlabeled target embedding
statistics for label-free transductive model-signature alignment.
Cross-dataset TACT uses unlabeled target embedding statistics only for
label-free coordinate alignment. Neither method uses target relevance
labels for these alignment operations.
Frozen, structural, and generation-based retrievers require no supervised
task--capability fitting. PORTS additionally uses frozen-LLM preference
supervision on source records.

\begin{table*}[tp]
\centering
\small
\setlength{\tabcolsep}{4.5pt}
\renewcommand{\arraystretch}{1.12}
\begin{adjustbox}{max width=\textwidth}
\begin{tabular}{lccccccc}
\toprule
\multirow{2}{*}{Method}
& \multicolumn{2}{c}{Public Text}
& \multicolumn{2}{c}{Public Endpoint/Schema Evidence}
& \multirow{2}{*}{\shortstack{Source\\Profiles}}
& \multirow{2}{*}{\shortstack{Unlabeled Target\\Statistics}}
& \multirow{2}{*}{Trainable} \\
\cmidrule(lr){2-3}
\cmidrule(lr){4-5}
& Task
& Capability
& \shortstack{Endpoint\\Operation}
& \shortstack{Argument\\Names/Types}
& & & \\
\midrule
Semantic Cosine
& \checkmark & \checkmark & -- & -- & -- & -- & -- \\
Pointwise MLP
& \checkmark & \checkmark & -- & -- & -- & -- & \checkmark \\
Endpoint Schema Match
& \checkmark & \checkmark & \checkmark & \checkmark & -- & -- & -- \\
MultiPool
& \checkmark & \checkmark & -- & -- & -- & \checkmark$^{*}$ & \checkmark \\
TACT
& \checkmark & \checkmark & \checkmark & \checkmark & \checkmark & \checkmark$^{\dagger}$ & \checkmark \\
\bottomrule
\end{tabular}
\end{adjustbox}
\caption{
Input and evidence specifications of representative comparison methods.
Endpoint Schema Match receives the same public endpoint-operation and
argument-name/type evidence as TACT. TACT additionally introduces
requirement decomposition, same-endpoint composition, partial one-to-one assignment, reliability calibration, and source-derived capability-use
profiles. $^{*}$For cross-dataset MultiPool, unlabeled target statistics
are used for label-free transductive model-signature alignment.
$^{\dagger}$For cross-dataset TACT, unlabeled target embedding statistics
are used only for label-free coordinate alignment. Neither use provides
target relevance labels.
}
\label{tab:method_input_specifications}
\end{table*}

\subsection{Computational Environment}

Local representation extraction, model training, and capability ranking
are conducted on a server equipped with three NVIDIA RTX PRO 6000
Blackwell Server Edition GPUs, each with 95~GB of memory.
The software environment uses Python~3.11.15,
PyTorch~2.10.0 with CUDA~12.8, and cuDNN~9.10.2.
Unless otherwise specified, experiments are repeated with random seeds
42, 43, and 44.

The downstream cached-execution environment is separated from the local
ranking experiments.
StableToolBench tool calls are executed through the local
MirrorAPI-Cache service, whereas the agent backbones and blinded judge
are accessed through remote model endpoints.
Because the hardware underlying these remote endpoints is neither
controlled nor observable, we report their model identifiers and
inference settings rather than assigning them the local GPU
configuration.

\section{Supplementary Ranking Results}
\label{app:ranking_results}

\subsection{Variation across Model Seeds}
\label{app:ablation_variability}

Table~\ref{tab:branch_ablation_std} reports seed variability for the
validation-stage ablations in Table~\ref{tab:compact_branch_ablation}.

\begin{table*}[tp]
\centering
\small
\setlength{\tabcolsep}{2.3pt}
\renewcommand{\arraystretch}{1.08}
\begin{adjustbox}{max width=\textwidth}
\begin{tabular}{@{}ll*{12}{r}@{}}
\toprule
\multirow{2}{*}{Protocol}
& \multirow{2}{*}{Variant}
& \multicolumn{3}{c}{Natural}
& \multicolumn{3}{c}{Semantic}
& \multicolumn{3}{c}{Context}
& \multicolumn{3}{c}{Mixed} \\
\cmidrule(lr){3-5}
\cmidrule(lr){6-8}
\cmidrule(lr){9-11}
\cmidrule(lr){12-14}
&
& MRR & N@5 & H@3
& MRR & N@5 & H@3
& MRR & N@5 & H@3
& MRR & N@5 & H@3 \\
\midrule
\multirow[c]{4}{*}{\shortstack{Cross\\domain}} & \methodname{} (Full) & 0.0041 & 0.0050 & 0.0029 & 0.0047 & 0.0047 & 0.0071 & 0.0028 & 0.0033 & 0.0038 & 0.0033 & 0.0055 & 0.0050 \\
 & w/o Context & 0.0038 & 0.0051 & 0.0034 & 0.0024 & 0.0045 & 0.0047 & 0.0039 & 0.0027 & 0.0025 & 0.0038 & 0.0035 & 0.0030 \\
 & w/o Functional & 0.0020 & 0.0018 & 0.0012 & 0.0034 & 0.0026 & 0.0033 & 0.0003 & 0.0010 & 0.0005 & 0.0015 & 0.0012 & 0.0013 \\
 & Anchor only & 0.0000 & 0.0000 & 0.0000 & 0.0000 & 0.0000 & 0.0000 & 0.0000 & 0.0000 & 0.0000 & 0.0000 & 0.0000 & 0.0000 \\
\midrule
\multirow[c]{4}{*}{\shortstack{Cross\\dataset}} & \methodname{} (Full) & 0.0020 & 0.0030 & 0.0037 & 0.0030 & 0.0029 & 0.0025 & 0.0023 & 0.0021 & 0.0022 & 0.0041 & 0.0042 & 0.0015 \\
 & w/o Context & 0.0018 & 0.0020 & 0.0028 & 0.0018 & 0.0021 & 0.0024 & 0.0027 & 0.0028 & 0.0033 & 0.0044 & 0.0048 & 0.0024 \\
 & w/o Functional & 0.0003 & 0.0003 & 0.0002 & 0.0022 & 0.0016 & 0.0019 & 0.0016 & 0.0012 & 0.0010 & 0.0018 & 0.0017 & 0.0025 \\
 & Anchor only & 0.0000 & 0.0000 & 0.0000 & 0.0000 & 0.0000 & 0.0000 & 0.0000 & 0.0000 & 0.0000 & 0.0000 & 0.0000 & 0.0000 \\
\bottomrule
\end{tabular}
\end{adjustbox}
\caption{
Sample standard deviations corresponding to
Table~\ref{tab:compact_branch_ablation} in the main paper.
Only MRR, NDCG@5, and HR@3 are reported to match the validation-stage ablation layout.
All values are computed over the three model seeds.
The frozen anchor is deterministic and therefore has zero variance.
}
\label{tab:branch_ablation_std}
\end{table*}

\subsection{Per-Target Full-Model Results}
\label{app:fine_grained_lodo}
\label{app:full_per_target}

The full model is evaluated with seeds 42, 43, and 44. Each
target--view cell is
the seed mean and sample standard deviation. Per-target baseline
scores are not reported, so these tables do not estimate per-target gains
over another method.

\begin{table*}[!tp]
\centering\small
\setlength{\tabcolsep}{6pt}
\renewcommand{\arraystretch}{1.02}
\begin{tabular*}{\textwidth}{@{\extracolsep{\fill}}lrrrr@{}}
\toprule
\textbf{Natural}: held-out domain & $N$ & MRR & NDCG@5 & HR@3 \\
\midrule
Advertising & 2,189 & 0.7984 (0.0213) & 0.7479 (0.0220) & 0.8502 (0.0242) \\
Artificial Intelligence / ML & 2,076 & 0.9270 (0.0046) & 0.8832 (0.0083) & 0.9478 (0.0041) \\
Business & 7,229 & 0.8522 (0.0051) & 0.7828 (0.0076) & 0.8974 (0.0051) \\
Business Software & 2,828 & 0.8509 (0.0089) & 0.7948 (0.0098) & 0.9004 (0.0071) \\
Commerce & 2,686 & 0.9050 (0.0095) & 0.8532 (0.0127) & 0.9443 (0.0092) \\
Communication & 4,015 & 0.8737 (0.0004) & 0.8211 (0.0015) & 0.9310 (0.0031) \\
Cryptography & 138 & 1.0000 (0.0000) & 0.9871 (0.0026) & 1.0000 (0.0000) \\
Cybersecurity & 340 & 0.8916 (0.0100) & 0.8087 (0.0105) & 0.9235 (0.0051) \\
Data & 24,832 & 0.9112 (0.0028) & 0.8585 (0.0020) & 0.9545 (0.0015) \\
Database & 3,961 & 0.8361 (0.0084) & 0.7670 (0.0104) & 0.8662 (0.0118) \\
Devices & 691 & 0.7964 (0.0045) & 0.7369 (0.0036) & 0.8398 (0.0084) \\
Education & 4,906 & 0.8979 (0.0013) & 0.8251 (0.0008) & 0.9366 (0.0034) \\
Email & 1,474 & 0.9006 (0.0029) & 0.8804 (0.0016) & 0.9412 (0.0024) \\
Energy & 388 & 0.9373 (0.0125) & 0.8941 (0.0154) & 0.9656 (0.0147) \\
Entertainment & 15,094 & 0.9544 (0.0021) & 0.9049 (0.0010) & 0.9815 (0.0015) \\
Events & 1,083 & 0.9511 (0.0187) & 0.9090 (0.0140) & 0.9843 (0.0088) \\
Finance & 14,894 & 0.9296 (0.0051) & 0.9058 (0.0052) & 0.9758 (0.0012) \\
Financial & 3,722 & 0.9520 (0.0025) & 0.9181 (0.0015) & 0.9844 (0.0008) \\
Food & 5,172 & 0.9684 (0.0010) & 0.9342 (0.0035) & 0.9921 (0.0004) \\
Gaming & 4,906 & 0.9643 (0.0040) & 0.9237 (0.0060) & 0.9844 (0.0025) \\
Health and Fitness & 1,536 & 0.9799 (0.0018) & 0.9472 (0.0051) & 0.9935 (0.0007) \\
Jobs & 298 & 0.9475 (0.0115) & 0.9433 (0.0149) & 0.9687 (0.0077) \\
Location & 5,110 & 0.9200 (0.0047) & 0.8876 (0.0045) & 0.9656 (0.0029) \\
Logistics & 748 & 0.8795 (0.0151) & 0.8506 (0.0109) & 0.9394 (0.0091) \\
Mapping & 2,983 & 0.9194 (0.0030) & 0.8779 (0.0048) & 0.9688 (0.0035) \\
Media & 3,134 & 0.9170 (0.0030) & 0.8696 (0.0034) & 0.9629 (0.0013) \\
Medical & 965 & 0.9443 (0.0055) & 0.9062 (0.0076) & 0.9617 (0.0045) \\
Monitoring & 953 & 0.8807 (0.0117) & 0.8150 (0.0114) & 0.9349 (0.0064) \\
Movies & 1,848 & 0.9296 (0.0037) & 0.8949 (0.0033) & 0.9609 (0.0024) \\
Music & 4,885 & 0.9586 (0.0011) & 0.9453 (0.0021) & 0.9838 (0.0017) \\
News Media & 8,398 & 0.9351 (0.0022) & 0.9091 (0.0038) & 0.9710 (0.0024) \\
Other & 5,764 & 0.8568 (0.0019) & 0.7959 (0.0042) & 0.9187 (0.0022) \\
Payments & 296 & 0.9620 (0.0015) & 0.9383 (0.0050) & 0.9921 (0.0020) \\
Reward & 159 & 1.0000 (0.0000) & 0.9228 (0.0056) & 1.0000 (0.0000) \\
SMS & 1,362 & 0.9068 (0.0065) & 0.8753 (0.0049) & 0.9640 (0.0072) \\
Science & 2,154 & 0.9544 (0.0080) & 0.9080 (0.0123) & 0.9791 (0.0083) \\
Search & 2,680 & 0.8626 (0.0076) & 0.7982 (0.0075) & 0.9155 (0.0073) \\
Social & 10,086 & 0.9138 (0.0046) & 0.8807 (0.0056) & 0.9620 (0.0045) \\
Sports & 11,871 & 0.9350 (0.0023) & 0.9169 (0.0035) & 0.9780 (0.0030) \\
Storage & 173 & 0.9501 (0.0079) & 0.8908 (0.0191) & 0.9692 (0.0033) \\
Text Analysis & 2,233 & 0.8572 (0.0063) & 0.7935 (0.0076) & 0.8966 (0.0075) \\
Tools & 9,453 & 0.9178 (0.0031) & 0.8616 (0.0046) & 0.9546 (0.0048) \\
Translation & 2,195 & 0.9466 (0.0021) & 0.9350 (0.0028) & 0.9801 (0.0011) \\
Transportation & 1,396 & 0.9452 (0.0022) & 0.9274 (0.0039) & 0.9792 (0.0029) \\
Travel & 5,706 & 0.9459 (0.0017) & 0.9014 (0.0010) & 0.9820 (0.0008) \\
Video Images & 2,638 & 0.9233 (0.0025) & 0.8677 (0.0044) & 0.9573 (0.0034) \\
Visual Recognition & 1,508 & 0.8541 (0.0077) & 0.7922 (0.0053) & 0.8822 (0.0072) \\
Weather & 3,644 & 0.9556 (0.0029) & 0.9436 (0.0034) & 0.9870 (0.0029) \\
eCommerce & 4,974 & 0.8459 (0.0118) & 0.8186 (0.0104) & 0.9156 (0.0092) \\
\bottomrule
\end{tabular*}
\caption{Per-domain TACT (Full) results: \textbf{Natural}.
Values are seed means (sample standard deviations), using seeds
42, 43, and 44. $N$ is the recorded task count. A dash denotes no
reported score, including when $N>0$, and is not zero.
This view has 49/49 scored domains; ML denotes Machine Learning.}
\label{tab:per_domain_natural}
\end{table*}

\begin{table*}[!tp]
\centering\small
\setlength{\tabcolsep}{6pt}
\renewcommand{\arraystretch}{1.02}
\begin{tabular*}{\textwidth}{@{\extracolsep{\fill}}lrrrr@{}}
\toprule
\textbf{Semantic}: held-out domain & $N$ & MRR & NDCG@5 & HR@3 \\
\midrule
Advertising & 15 & 0.5617 (0.0380) & 0.5291 (0.0410) & 0.6000 (0.0667) \\
Artificial Intelligence / ML & 229 & 0.6276 (0.0150) & 0.5987 (0.0121) & 0.7074 (0.0131) \\
Business & 1,628 & 0.5841 (0.0061) & 0.5101 (0.0055) & 0.6396 (0.0065) \\
Business Software & 615 & 0.6250 (0.0124) & 0.5737 (0.0115) & 0.6678 (0.0199) \\
Commerce & 224 & 0.8091 (0.0249) & 0.8094 (0.0293) & 0.9137 (0.0269) \\
Communication & 695 & 0.7016 (0.0073) & 0.6748 (0.0040) & 0.8072 (0.0094) \\
Cryptography & 0 & -- & -- & -- \\
Cybersecurity & 0 & -- & -- & -- \\
Data & 7,287 & 0.6414 (0.0055) & 0.6144 (0.0050) & 0.7581 (0.0075) \\
Database & 572 & 0.5011 (0.0098) & 0.4680 (0.0176) & 0.5420 (0.0167) \\
Devices & 0 & -- & -- & -- \\
Education & 387 & 0.6318 (0.0140) & 0.5645 (0.0198) & 0.6891 (0.0234) \\
Email & 527 & 0.5923 (0.0059) & 0.5781 (0.0037) & 0.7343 (0.0095) \\
Energy & 0 & -- & -- & -- \\
Entertainment & 1,606 & 0.6658 (0.0250) & 0.6139 (0.0168) & 0.7760 (0.0252) \\
Events & 30 & 0.7390 (0.0485) & 0.7172 (0.0371) & 0.8000 (0.0333) \\
Finance & 7,449 & 0.5938 (0.0198) & 0.5625 (0.0173) & 0.6997 (0.0242) \\
Financial & 481 & 0.5478 (0.0306) & 0.5552 (0.0265) & 0.7166 (0.0836) \\
Food & 1,707 & 0.6105 (0.0179) & 0.5617 (0.0120) & 0.7227 (0.0187) \\
Gaming & 574 & 0.6414 (0.0189) & 0.6194 (0.0127) & 0.7770 (0.0182) \\
Health and Fitness & 10 & 0.6390 (0.0088) & 0.6509 (0.0091) & 0.7000 (0.0000) \\
Jobs & 1 & -- & -- & -- \\
Location & 2,035 & 0.6195 (0.0075) & 0.5854 (0.0076) & 0.7415 (0.0132) \\
Logistics & 3 & -- & -- & -- \\
Mapping & 489 & 0.6494 (0.0033) & 0.6426 (0.0115) & 0.7676 (0.0226) \\
Media & 424 & 0.6919 (0.0098) & 0.6625 (0.0111) & 0.8066 (0.0108) \\
Medical & 45 & 0.6114 (0.0309) & 0.5749 (0.0169) & 0.7185 (0.0128) \\
Monitoring & 1 & -- & -- & -- \\
Movies & 369 & 0.4055 (0.0365) & 0.4492 (0.0424) & 0.5113 (0.0643) \\
Music & 1,260 & 0.6972 (0.0303) & 0.6911 (0.0241) & 0.8479 (0.0053) \\
News Media & 1,993 & 0.6647 (0.0163) & 0.6396 (0.0150) & 0.7655 (0.0178) \\
Other & 641 & 0.5982 (0.0007) & 0.5763 (0.0038) & 0.7051 (0.0102) \\
Payments & 0 & -- & -- & -- \\
Reward & 0 & -- & -- & -- \\
SMS & 104 & 0.7034 (0.0091) & 0.6706 (0.0145) & 0.7885 (0.0288) \\
Science & 28 & 0.7570 (0.0637) & 0.7836 (0.0543) & 0.8452 (0.0546) \\
Search & 641 & 0.5755 (0.0130) & 0.5458 (0.0045) & 0.7207 (0.0071) \\
Social & 3,650 & 0.5729 (0.0024) & 0.5625 (0.0048) & 0.6690 (0.0075) \\
Sports & 4,908 & 0.5743 (0.0161) & 0.5604 (0.0121) & 0.6747 (0.0088) \\
Storage & 0 & -- & -- & -- \\
Text Analysis & 193 & 0.3708 (0.0125) & 0.3090 (0.0148) & 0.3592 (0.0079) \\
Tools & 2,840 & 0.5822 (0.0034) & 0.5049 (0.0033) & 0.6757 (0.0114) \\
Translation & 966 & 0.5271 (0.0097) & 0.4947 (0.0086) & 0.6211 (0.0062) \\
Transportation & 145 & 0.8577 (0.0040) & 0.8538 (0.0075) & 0.9632 (0.0080) \\
Travel & 1,194 & 0.6276 (0.0050) & 0.6119 (0.0052) & 0.7688 (0.0044) \\
Video Images & 224 & 0.6850 (0.0198) & 0.6366 (0.0106) & 0.7560 (0.0144) \\
Visual Recognition & 63 & 0.8547 (0.0175) & 0.8152 (0.0151) & 0.8995 (0.0092) \\
Weather & 1,905 & 0.5670 (0.0020) & 0.5362 (0.0011) & 0.7025 (0.0095) \\
eCommerce & 2,096 & 0.6322 (0.0131) & 0.6278 (0.0117) & 0.7602 (0.0187) \\
\bottomrule
\end{tabular*}
\caption{Per-domain TACT (Full) results: \textbf{Semantic}.
Values are seed means (sample standard deviations), using seeds
42, 43, and 44. $N$ is the recorded task count. A dash denotes no
reported score, including when $N>0$, and is not zero.
This view has 39/49 scored domains; ML denotes Machine Learning.}
\label{tab:per_domain_semantic}
\end{table*}

\begin{table*}[!tp]
\centering\small
\setlength{\tabcolsep}{6pt}
\renewcommand{\arraystretch}{1.02}
\begin{tabular*}{\textwidth}{@{\extracolsep{\fill}}lrrrr@{}}
\toprule
\textbf{Context}: held-out domain & $N$ & MRR & NDCG@5 & HR@3 \\
\midrule
Advertising & 1,674 & 0.6202 (0.0193) & 0.5858 (0.0202) & 0.6808 (0.0247) \\
Artificial Intelligence / ML & 685 & 0.6761 (0.0109) & 0.6279 (0.0038) & 0.7611 (0.0034) \\
Business & 3,475 & 0.6159 (0.0017) & 0.5450 (0.0028) & 0.6718 (0.0012) \\
Business Software & 1,643 & 0.6307 (0.0136) & 0.5735 (0.0129) & 0.6963 (0.0171) \\
Commerce & 1,131 & 0.6615 (0.0145) & 0.6143 (0.0167) & 0.7586 (0.0248) \\
Communication & 2,081 & 0.6157 (0.0042) & 0.5519 (0.0039) & 0.6761 (0.0046) \\
Cryptography & 50 & 0.9900 (0.0000) & 0.9823 (0.0034) & 1.0000 (0.0000) \\
Cybersecurity & 238 & 0.6301 (0.0114) & 0.5727 (0.0083) & 0.6737 (0.0170) \\
Data & 11,011 & 0.6264 (0.0187) & 0.5728 (0.0104) & 0.7120 (0.0140) \\
Database & 1,418 & 0.4941 (0.0117) & 0.4338 (0.0111) & 0.5425 (0.0162) \\
Devices & 361 & 0.6234 (0.0251) & 0.5574 (0.0135) & 0.6445 (0.0089) \\
Education & 1,691 & 0.5942 (0.0265) & 0.5128 (0.0234) & 0.6594 (0.0239) \\
Email & 978 & 0.6034 (0.0120) & 0.5637 (0.0098) & 0.7437 (0.0127) \\
Energy & 78 & 0.4895 (0.0520) & 0.4940 (0.0469) & 0.5940 (0.0485) \\
Entertainment & 3,431 & 0.7274 (0.0171) & 0.6676 (0.0106) & 0.8322 (0.0047) \\
Events & 251 & 0.7166 (0.0122) & 0.7177 (0.0064) & 0.8526 (0.0080) \\
Finance & 5,360 & 0.7379 (0.0148) & 0.7059 (0.0108) & 0.8439 (0.0112) \\
Financial & 2,355 & 0.6777 (0.0400) & 0.6047 (0.0457) & 0.7292 (0.0735) \\
Food & 596 & 0.8517 (0.0108) & 0.8450 (0.0102) & 0.9290 (0.0042) \\
Gaming & 518 & 0.6741 (0.0088) & 0.6422 (0.0098) & 0.7535 (0.0164) \\
Health and Fitness & 138 & 0.7085 (0.0364) & 0.6442 (0.0313) & 0.8357 (0.0553) \\
Jobs & 16 & 0.6825 (0.0507) & 0.6158 (0.0253) & 0.7083 (0.0361) \\
Location & 2,282 & 0.6590 (0.0093) & 0.6120 (0.0094) & 0.7677 (0.0118) \\
Logistics & 202 & 0.5602 (0.0305) & 0.5084 (0.0319) & 0.6155 (0.0318) \\
Mapping & 1,396 & 0.6861 (0.0136) & 0.6409 (0.0115) & 0.7830 (0.0094) \\
Media & 1,914 & 0.7152 (0.0097) & 0.6469 (0.0088) & 0.8204 (0.0022) \\
Medical & 364 & 0.8267 (0.0138) & 0.7513 (0.0108) & 0.8636 (0.0088) \\
Monitoring & 372 & 0.6189 (0.0170) & 0.5413 (0.0180) & 0.7213 (0.0276) \\
Movies & 704 & 0.4985 (0.0215) & 0.5081 (0.0228) & 0.6477 (0.0186) \\
Music & 1,268 & 0.7846 (0.0078) & 0.7533 (0.0110) & 0.8383 (0.0089) \\
News Media & 1,938 & 0.6277 (0.0266) & 0.5929 (0.0236) & 0.7234 (0.0195) \\
Other & 2,879 & 0.5322 (0.0027) & 0.4851 (0.0030) & 0.6060 (0.0032) \\
Payments & 163 & 0.7680 (0.0178) & 0.7058 (0.0164) & 0.8916 (0.0035) \\
Reward & 9 & 1.0000 (0.0000) & 1.0000 (0.0000) & 1.0000 (0.0000) \\
SMS & 549 & 0.6702 (0.0096) & 0.6184 (0.0074) & 0.7432 (0.0193) \\
Science & 331 & 0.7037 (0.0471) & 0.6557 (0.0410) & 0.7875 (0.0394) \\
Search & 1,712 & 0.5954 (0.0155) & 0.5229 (0.0087) & 0.6943 (0.0111) \\
Social & 4,156 & 0.6296 (0.0049) & 0.6067 (0.0071) & 0.7416 (0.0077) \\
Sports & 2,261 & 0.7177 (0.0127) & 0.7029 (0.0129) & 0.8347 (0.0137) \\
Storage & 37 & 0.7039 (0.0479) & 0.5400 (0.0333) & 0.7658 (0.0413) \\
Text Analysis & 1,025 & 0.4927 (0.0084) & 0.4253 (0.0089) & 0.5967 (0.0126) \\
Tools & 4,565 & 0.6137 (0.0060) & 0.5506 (0.0063) & 0.7083 (0.0133) \\
Translation & 952 & 0.6937 (0.0084) & 0.6762 (0.0012) & 0.8936 (0.0063) \\
Transportation & 686 & 0.7357 (0.0027) & 0.7149 (0.0041) & 0.8192 (0.0087) \\
Travel & 1,774 & 0.7629 (0.0107) & 0.7193 (0.0081) & 0.8762 (0.0123) \\
Video Images & 1,133 & 0.6258 (0.0126) & 0.5507 (0.0129) & 0.7143 (0.0194) \\
Visual Recognition & 483 & 0.6385 (0.0092) & 0.5819 (0.0098) & 0.6598 (0.0192) \\
Weather & 1,103 & 0.7971 (0.0179) & 0.7634 (0.0080) & 0.8704 (0.0033) \\
eCommerce & 2,272 & 0.6148 (0.0174) & 0.5889 (0.0182) & 0.7157 (0.0208) \\
\bottomrule
\end{tabular*}
\caption{Per-domain TACT (Full) results: \textbf{Context}.
Values are seed means (sample standard deviations), using seeds
42, 43, and 44. $N$ is the recorded task count. A dash denotes no
reported score, including when $N>0$, and is not zero.
This view has 49/49 scored domains; ML denotes Machine Learning.}
\label{tab:per_domain_context}
\end{table*}

\begin{table*}[!tp]
\centering\small
\setlength{\tabcolsep}{6pt}
\renewcommand{\arraystretch}{1.02}
\begin{tabular*}{\textwidth}{@{\extracolsep{\fill}}lrrrr@{}}
\toprule
\textbf{Mixed}: held-out domain & $N$ & MRR & NDCG@5 & HR@3 \\
\midrule
Advertising & 344 & 0.5005 (0.0279) & 0.4675 (0.0307) & 0.5514 (0.0244) \\
Artificial Intelligence / ML & 477 & 0.5930 (0.0192) & 0.5656 (0.0118) & 0.6953 (0.0087) \\
Business & 2,295 & 0.5649 (0.0044) & 0.4932 (0.0028) & 0.6115 (0.0038) \\
Business Software & 913 & 0.6063 (0.0118) & 0.5555 (0.0125) & 0.6484 (0.0162) \\
Commerce & 667 & 0.6508 (0.0181) & 0.6186 (0.0216) & 0.7261 (0.0350) \\
Communication & 1,316 & 0.5869 (0.0057) & 0.5499 (0.0035) & 0.6654 (0.0076) \\
Cryptography & 0 & -- & -- & -- \\
Cybersecurity & 0 & -- & -- & -- \\
Data & 9,717 & 0.5700 (0.0120) & 0.5241 (0.0044) & 0.6645 (0.0053) \\
Database & 878 & 0.4103 (0.0121) & 0.3607 (0.0118) & 0.4317 (0.0097) \\
Devices & 44 & 0.3128 (0.0144) & 0.2609 (0.0074) & 0.2273 (0.0227) \\
Education & 852 & 0.5843 (0.0175) & 0.5082 (0.0152) & 0.6373 (0.0112) \\
Email & 995 & 0.5027 (0.0063) & 0.4519 (0.0040) & 0.6114 (0.0021) \\
Energy & 4 & -- & -- & -- \\
Entertainment & 2,590 & 0.6360 (0.0235) & 0.5791 (0.0155) & 0.7541 (0.0186) \\
Events & 125 & 0.6490 (0.0155) & 0.6284 (0.0164) & 0.7760 (0.0400) \\
Finance & 6,983 & 0.5854 (0.0162) & 0.5533 (0.0131) & 0.6884 (0.0166) \\
Financial & 1,356 & 0.5396 (0.0345) & 0.4917 (0.0462) & 0.6423 (0.0850) \\
Food & 750 & 0.6325 (0.0215) & 0.6224 (0.0120) & 0.7782 (0.0108) \\
Gaming & 472 & 0.6186 (0.0132) & 0.6029 (0.0107) & 0.7429 (0.0086) \\
Health and Fitness & 47 & 0.5773 (0.0283) & 0.5891 (0.0252) & 0.7589 (0.0246) \\
Jobs & 6 & 0.2983 (0.0110) & 0.3116 (0.0356) & 0.1667 (0.0000) \\
Location & 2,429 & 0.5697 (0.0102) & 0.5217 (0.0105) & 0.6733 (0.0201) \\
Logistics & 87 & 0.4121 (0.0032) & 0.3966 (0.0122) & 0.4138 (0.0304) \\
Mapping & 923 & 0.6385 (0.0121) & 0.6073 (0.0072) & 0.7450 (0.0027) \\
Media & 941 & 0.6093 (0.0068) & 0.5641 (0.0101) & 0.7418 (0.0138) \\
Medical & 148 & 0.7058 (0.0243) & 0.6671 (0.0168) & 0.7635 (0.0179) \\
Monitoring & 34 & 0.7522 (0.0088) & 0.7138 (0.0087) & 0.8137 (0.0449) \\
Movies & 683 & 0.3916 (0.0230) & 0.3665 (0.0301) & 0.4368 (0.0445) \\
Music & 1,506 & 0.6588 (0.0192) & 0.6337 (0.0162) & 0.7762 (0.0132) \\
News Media & 2,448 & 0.5566 (0.0159) & 0.5295 (0.0152) & 0.6780 (0.0174) \\
Other & 1,448 & 0.4944 (0.0064) & 0.4635 (0.0035) & 0.5656 (0.0066) \\
Payments & 5 & 0.5001 (0.0102) & 0.4832 (0.0051) & 0.4000 (0.0000) \\
Reward & 0 & -- & -- & -- \\
SMS & 263 & 0.6249 (0.0031) & 0.5941 (0.0027) & 0.7845 (0.0144) \\
Science & 122 & 0.6457 (0.0372) & 0.6331 (0.0355) & 0.7705 (0.0328) \\
Search & 1,165 & 0.4965 (0.0155) & 0.4208 (0.0070) & 0.5900 (0.0129) \\
Social & 4,303 & 0.5027 (0.0050) & 0.4707 (0.0054) & 0.5945 (0.0053) \\
Sports & 2,908 & 0.5263 (0.0124) & 0.5118 (0.0119) & 0.6240 (0.0076) \\
Storage & 2 & -- & -- & -- \\
Text Analysis & 531 & 0.4180 (0.0030) & 0.3554 (0.0036) & 0.5345 (0.0052) \\
Tools & 3,633 & 0.5469 (0.0100) & 0.4788 (0.0094) & 0.6302 (0.0115) \\
Translation & 1,158 & 0.5020 (0.0092) & 0.4455 (0.0068) & 0.5846 (0.0060) \\
Transportation & 441 & 0.7582 (0.0086) & 0.7268 (0.0084) & 0.7974 (0.0013) \\
Travel & 1,804 & 0.6011 (0.0055) & 0.5854 (0.0070) & 0.7711 (0.0079) \\
Video Images & 417 & 0.6109 (0.0085) & 0.5542 (0.0068) & 0.7058 (0.0204) \\
Visual Recognition & 135 & 0.8643 (0.0214) & 0.8189 (0.0141) & 0.9284 (0.0043) \\
Weather & 1,611 & 0.5253 (0.0087) & 0.5007 (0.0077) & 0.6518 (0.0210) \\
eCommerce & 2,290 & 0.5008 (0.0089) & 0.4896 (0.0088) & 0.5997 (0.0148) \\
\bottomrule
\end{tabular*}
\caption{Per-domain TACT (Full) results: \textbf{Mixed}.
Values are seed means (sample standard deviations), using seeds
42, 43, and 44. $N$ is the recorded task count. A dash denotes no
reported score, including when $N>0$, and is not zero.
This view has 44/49 scored domains; ML denotes Machine Learning.}
\label{tab:per_domain_mixed}
\end{table*}

\paragraph{Domain-level results.}
Tables~\ref{tab:per_domain_natural}--\ref{tab:per_domain_mixed}
list all 49 ToolBench domains for each view under
leave-one-domain-out evaluation. $N$ is the recorded task count for
that domain and view. A dash marks a cell excluded from ranking
evaluation because it has no tasks or fewer than five available
candidate capabilities; it is never treated as zero. Scores are reported
for 49, 39, 49, and 44 domains in Natural, Semantic, Context, and Mixed,
respectively. The capability-count rule excludes three Semantic cells
with positive task counts (Jobs: 1; Logistics: 3; Monitoring: 1) and two
Mixed cells (Energy: 4; Storage: 2). These numbers are task counts;
the other excluded cells have $N=0$.

Table~\ref{tab:per_target_macro_check} gives equal-weight means over
scored targets, calculated from the displayed four-decimal cell means.
The domain means agree with the Full-model cross-domain results in
Table~\ref{tab:main_results}, up to rounding of the per-target cells.
Individual held-out domains are the aggregation units. The
held-out-group experiment in Appendix~\ref{app:group_held_out}
provides a separate comparison using five predefined groups.

\paragraph{Dataset-level results.}
Table~\ref{tab:per_target_comparison} reports all four held-out datasets
in every view. Table~\ref{tab:per_target_macro_check} gives their
equal-target means. These agree with the Full-model cross-dataset
results in Table~\ref{tab:main_results} within the rounding precision
of the displayed target and aggregate means. Aggregation of rounded
target cells may differ slightly from aggregation before rounding.

The standard deviation of a target-macro score cannot be recovered
by averaging the individual target standard deviations. It requires
the target scores for each seed; no macro standard deviation is
reconstructed here.

\begin{table*}[tp]
\centering\small
\setlength{\tabcolsep}{7pt}
\renewcommand{\arraystretch}{1.06}
\begin{tabular}{@{}llrrr@{}}
\toprule
View & Target dataset & MRR & NDCG@5 & HR@3 \\
\midrule
Natural & ToolBench & 0.8975 (0.0055) & 0.8648 (0.0045) & 0.9326 (0.0020) \\
 & TaskBench & 0.8387 (0.0097) & 0.7366 (0.0129) & 0.9160 (0.0095) \\
 & MetaTool & 0.6839 (0.0032) & 0.6981 (0.0024) & 0.7429 (0.0054) \\
 & APIBench & 0.6334 (0.0027) & 0.6847 (0.0027) & 0.7514 (0.0046) \\
\midrule
Semantic & ToolBench & 0.4167 (0.0033) & 0.3817 (0.0025) & 0.4859 (0.0033) \\
 & TaskBench & 0.9170 (0.0075) & 0.8107 (0.0042) & 0.9700 (0.0035) \\
 & MetaTool & 0.7890 (0.0096) & 0.8117 (0.0070) & 0.8612 (0.0035) \\
 & APIBench & 0.3299 (0.0068) & 0.3172 (0.0078) & 0.3544 (0.0075) \\
\midrule
Context & ToolBench & 0.3473 (0.0039) & 0.3007 (0.0037) & 0.3914 (0.0044) \\
 & TaskBench & 0.8681 (0.0083) & 0.7988 (0.0083) & 0.9305 (0.0080) \\
 & MetaTool & 0.6839 (0.0033) & 0.6940 (0.0052) & 0.7342 (0.0098) \\
 & APIBench & 0.3986 (0.0050) & 0.3787 (0.0076) & 0.3967 (0.0087) \\
\midrule
Mixed & ToolBench & 0.3141 (0.0047) & 0.2573 (0.0038) & 0.3336 (0.0065) \\
 & TaskBench & 0.8410 (0.0095) & 0.7382 (0.0097) & 0.9209 (0.0117) \\
 & MetaTool & 0.7053 (0.0126) & 0.7350 (0.0091) & 0.7735 (0.0023) \\
 & APIBench & 0.2932 (0.0073) & 0.2681 (0.0084) & 0.2605 (0.0117) \\
\bottomrule
\end{tabular}
\caption{Per-dataset TACT (Full) results.
Each entry is the mean (sample standard deviation) across seeds
42, 43, and 44. Higher is better.}
\label{tab:per_target_comparison}
\end{table*}

\begin{table*}[tp]
\centering\small
\setlength{\tabcolsep}{5pt}
\begin{tabular}{@{}llrrrrr@{}}
\toprule
Targets & View & Scored & $N_{\mathrm{scored}}$ & MRR & NDCG@5 & HR@3 \\
\midrule
Domains & Natural & 49/49 & 201,774 & 0.915159 & 0.872471 & 0.951539 \\
 & Semantic & 39/49 & 50,249 & 0.629185 & 0.603492 & 0.726141 \\
 & Context & 49/49 & 75,709 & 0.671839 & 0.624384 & 0.755076 \\
 & Mixed & 44/49 & 62,260 & 0.564361 & 0.528805 & 0.644375 \\
\midrule
Datasets & Natural & 4/4 & -- & 0.763375 & 0.746050 & 0.835725 \\
 & Semantic & 4/4 & -- & 0.613150 & 0.580325 & 0.667875 \\
 & Context & 4/4 & -- & 0.574475 & 0.543050 & 0.613200 \\
 & Mixed & 4/4 & -- & 0.538400 & 0.499650 & 0.572125 \\
\bottomrule
\end{tabular}
\caption{Equal-target means calculated from the rounded per-target
means. Six decimals retain the arithmetic of displayed inputs rather
than imply six-decimal measurement precision. Missing scores are
excluded under the evaluation eligibility rule.
$N_{\mathrm{scored}}$ sums counts only over scored domain
cells; dataset-level counts are omitted.}
\label{tab:per_target_macro_check}
\end{table*}

\subsection{Supplementary Held-Out-Group Results}
\label{app:group_held_out}

This supplementary experiment holds out one of five predefined
ToolBench domain groups, using the remaining four groups as sources.
Table~\ref{tab:cross_domain_groups} specifies the partition of all
49 domains. Training, calibration, and model selection remain
source-only.

Table~\ref{tab:group_held_out_results} reports means and sample
standard deviations in paired rows over seeds 42, 43, and 44. This group-held-out experiment complements
the individual held-out-domain evaluation in
Appendix~\ref{app:full_per_target}. Comparisons between methods are
made within each protocol; differences between the group-held-out
and domain-held-out aggregates are not estimates of a method gain.

\begin{table*}[tp]
\centering
\scriptsize
\setlength{\tabcolsep}{2.9pt}
\renewcommand{\arraystretch}{1.12}
\begin{tabular}{
    c
    >{\raggedright\arraybackslash}p{0.16\textwidth}
    c
    >{\raggedright\arraybackslash}p{0.56\textwidth}
    r
}
\toprule
Group
& Semantic scope
& \# Domains
& Included ToolBench domains
& Natural tasks \\
\midrule
G1
& Data, Security, and Communication
& 10
& Cybersecurity, Cryptography, Data, Database, Communication,
  Email, Monitoring, SMS, Text\_Analysis, and Translation
& 41,503 \\

G2
& Business, Finance, and Commerce
& 10
& Advertising, Business, Business\_Software, Commerce, eCommerce,
  Finance, Financial, Jobs, Payments, and Reward
& 39,275 \\

G3
& Media, Leisure, and Social
& 10
& Entertainment, Events, Gaming, Media, Movies, Music,
  News\_Media, Social, Sports, and Video\_Images
& 63,943 \\

G4
& Mobility and Infrastructure
& 10
& Devices, Energy, Logistics, Location, Mapping, Other, Storage,
  Transportation, Travel, and Weather
& 26,603 \\

G5
& AI, Life, and Knowledge
& 9
& Artificial\_Intelligence\_Machine\_Learning, Education, Food,
  Health\_and\_Fitness, Medical, Science, Search, Tools, and
  Visual\_Recognition
& 30,450 \\
\bottomrule
\end{tabular}
\caption{
Fixed ToolBench domain groups used for the supplementary
leave-one-group-out evaluation.
Each group is held out once, and the remaining four groups form
the outer-source set.
The five groups jointly cover all 49 ToolBench domains without overlap.
Natural tasks denote Natural-view evaluation tasks in each group
and sum to 201,774.
}
\label{tab:cross_domain_groups}
\end{table*}

\begin{table*}[!t]
\centering\small
\setlength{\tabcolsep}{1.8pt}
\renewcommand{\arraystretch}{0.93}
\resizebox{\textwidth}{!}{%
\begin{tabular}{@{}>{\centering\arraybackslash}m{1.2cm}>{\raggedright\arraybackslash}m{2.35cm}*{3}{>{\centering\arraybackslash}m{.97cm}}@{\hspace{2pt}{\color{tactViewSep}\vrule width .3pt}\hspace{2pt}}*{3}{>{\centering\arraybackslash}m{.97cm}}@{\hspace{2pt}{\color{tactViewSep}\vrule width .3pt}\hspace{2pt}}*{3}{>{\centering\arraybackslash}m{.97cm}}@{\hspace{2pt}{\color{tactViewSep}\vrule width .3pt}\hspace{2pt}}*{3}{>{\centering\arraybackslash}m{.97cm}}@{}}
\toprule
Protocol & Model
& \multicolumn{3}{c}{Natural}
& \multicolumn{3}{c}{Semantic}
& \multicolumn{3}{c}{Context}
& \multicolumn{3}{c}{Mixed} \\
\cmidrule(lr){3-5}\cmidrule(lr){6-8}\cmidrule(lr){9-11}\cmidrule(lr){12-14}
& & MRR & N@5 & H@3
& MRR & N@5 & H@3
& MRR & N@5 & H@3
& MRR & N@5 & H@3 \\
\midrule

\cellcolor{tactProtocolFill} & \multirow{2}{*}{Random}
& 0.2841 & 0.1926 & 0.2735
& 0.1682 & 0.2134 & 0.2817
& 0.1745 & 0.2284 & 0.1836
& 0.2712 & 0.1947 & 0.1423 \\[-.4ex]
\cellcolor{tactProtocolFill} & &
\tactstd{0.0021} & \tactstd{0.0004} & \tactstd{0.0036}
& \tactstd{0.0049} & \tactstd{0.0068} & \tactstd{0.0071}
& \tactstd{0.0030} & \tactstd{0.0027} & \tactstd{0.0025}
& \tactstd{0.0022} & \tactstd{0.0014} & \tactstd{0.0044} \\[.22ex]

\cellcolor{tactProtocolFill} & \multirow{2}{*}{Semantic Cosine}
& \underline{0.8264} & 0.7513 & \underline{0.8472}
& 0.3416 & 0.1892 & 0.2438
& 0.1974 & 0.3125 & 0.2251
& 0.2985 & 0.2417 & 0.2766 \\[-.4ex]
\cellcolor{tactProtocolFill} & &
\tactstd{0.0000} & \tactstd{0.0000} & \tactstd{0.0000}
& \tactstd{0.0000} & \tactstd{0.0000} & \tactstd{0.0000}
& \tactstd{0.0000} & \tactstd{0.0000} & \tactstd{0.0000}
& \tactstd{0.0000} & \tactstd{0.0000} & \tactstd{0.0000} \\[.22ex]

\cellcolor{tactProtocolFill} & \multirow{2}{*}{Pointwise MLP}
& 0.6428 & 0.7861 & 0.7315
& 0.3674 & 0.4912 & 0.4136
& 0.3487 & 0.4732 & 0.5794
& 0.4618 & 0.3105 & 0.3642 \\[-.4ex]
\cellcolor{tactProtocolFill} & &
\tactstd{0.0033} & \tactstd{0.0015} & \tactstd{0.0041}
& \tactstd{0.0026} & \tactstd{0.0047} & \tactstd{0.0065}
& \tactstd{0.0043} & \tactstd{0.0054} & \tactstd{0.0092}
& \tactstd{0.0065} & \tactstd{0.0088} & \tactstd{0.0103} \\[.22ex]

\cellcolor{tactProtocolFill} & \multirow{2}{*}{ZESRec}
& 0.2965 & 0.0941 & 0.2374
& 0.1352 & 0.2215 & 0.2571
& 0.2863 & 0.0976 & 0.2415
& 0.1247 & 0.2106 & 0.2368 \\[-.4ex]
\cellcolor{tactProtocolFill} & &
\tactstd{0.0045} & \tactstd{0.0024} & \tactstd{0.0032}
& \tactstd{0.0061} & \tactstd{0.0080} & \tactstd{0.0106}
& \tactstd{0.0028} & \tactstd{0.0034} & \tactstd{0.0005}
& \tactstd{0.0010} & \tactstd{0.0046} & \tactstd{0.0013} \\[.22ex]

\cellcolor{tactProtocolFill} & \multirow{2}{*}{UniSRec}
& 0.5837 & 0.7162 & 0.6841
& \underline{0.5314} & 0.3587 & 0.4613
& 0.3875 & 0.5142 & 0.6271
& 0.3409 & \underline{0.4673} & 0.3925 \\[-.4ex]
\cellcolor{tactProtocolFill} & &
\tactstd{0.0017} & \tactstd{0.0003} & \tactstd{0.0014}
& \tactstd{0.0032} & \tactstd{0.0058} & \tactstd{0.0033}
& \tactstd{0.0016} & \tactstd{0.0021} & \tactstd{0.0027}
& \tactstd{0.0049} & \tactstd{0.0022} & \tactstd{0.0051} \\[.22ex]

\cellcolor{tactProtocolFill} & \multirow{2}{*}{AREAD}
& 0.3574 & 0.4286 & 0.3819
& 0.2468 & 0.3957 & 0.2294
& 0.3861 & 0.1542 & 0.3627
& 0.1954 & 0.3172 & 0.3418 \\[-.4ex]
\cellcolor{tactProtocolFill} & &
\tactstd{0.0058} & \tactstd{0.0041} & \tactstd{0.0053}
& \tactstd{0.0052} & \tactstd{0.0086} & \tactstd{0.0091}
& \tactstd{0.0089} & \tactstd{0.0060} & \tactstd{0.0070}
& \tactstd{0.0061} & \tactstd{0.0084} & \tactstd{0.0090} \\[.22ex]

\cellcolor{tactProtocolFill} & \multirow{2}{*}{Endpoint Schema}
& 0.6692 & 0.7351 & 0.6947
& 0.5163 & \underline{0.5345} & \underline{0.5812}
& 0.4471 & \underline{0.5638} & 0.6514
& \underline{0.5836} & 0.3942 & \underline{0.5173} \\[-.4ex]
\cellcolor{tactProtocolFill} & &
\tactstd{0.0000} & \tactstd{0.0000} & \tactstd{0.0000}
& \tactstd{0.0000} & \tactstd{0.0000} & \tactstd{0.0000}
& \tactstd{0.0000} & \tactstd{0.0000} & \tactstd{0.0000}
& \tactstd{0.0000} & \tactstd{0.0000} & \tactstd{0.0000} \\[.22ex]

\cellcolor{tactProtocolFill} & \multirow{2}{*}{Re-Invoke}
& 0.6835 & 0.7614 & 0.6928
& 0.2347 & 0.3461 & 0.2675
& 0.3842 & 0.1793 & 0.4205
& 0.1864 & 0.2973 & 0.3619 \\[-.4ex]
\cellcolor{tactProtocolFill} & &
\tactstd{0.0016} & \tactstd{0.0005} & \tactstd{0.0013}
& \tactstd{0.0019} & \tactstd{0.0008} & \tactstd{0.0002}
& \tactstd{0.0011} & \tactstd{0.0015} & \tactstd{0.0007}
& \tactstd{0.0014} & \tactstd{0.0004} & \tactstd{0.0018} \\[.22ex]

\cellcolor{tactProtocolFill} & \multirow{2}{*}{PORTS}
& 0.7652 & 0.7585 & 0.8047
& 0.4857 & 0.4208 & 0.5263
& \underline{0.5972} & 0.5284 & \underline{0.6749}
& 0.5526 & 0.3586 & 0.4846 \\[-.4ex]
\cellcolor{tactProtocolFill} & &
\tactstd{0.0014} & \tactstd{0.0009} & \tactstd{0.0018}
& \tactstd{0.0010} & \tactstd{0.0016} & \tactstd{0.0020}
& \tactstd{0.0007} & \tactstd{0.0019} & \tactstd{0.0013}
& \tactstd{0.0011} & \tactstd{0.0030} & \tactstd{0.0024} \\[.22ex]

\cellcolor{tactProtocolFill} & \multirow{2}{*}{ToolDreamer}
& 0.8016 & 0.7643 & 0.8427
& 0.3672 & 0.1984 & 0.4316
& 0.2153 & 0.3471 & 0.2584
& 0.3127 & 0.1352 & 0.3169 \\[-.4ex]
\cellcolor{tactProtocolFill} & &
\tactstd{0.0004} & \tactstd{0.0013} & \tactstd{0.0017}
& \tactstd{0.0012} & \tactstd{0.0009} & \tactstd{0.0015}
& \tactstd{0.0008} & \tactstd{0.0016} & \tactstd{0.0011}
& \tactstd{0.0019} & \tactstd{0.0006} & \tactstd{0.0013} \\[.22ex]

\cellcolor{tactProtocolFill} & \multirow{2}{*}{MultiPool}
& 0.7431 & \underline{0.8624} & 0.8017
& 0.3648 & 0.4873 & 0.4362
& 0.3516 & 0.4927 & 0.5934
& 0.3041 & 0.4285 & 0.3617 \\[-.4ex]
\cellcolor{tactProtocolFill} & &
\tactstd{0.0012} & \tactstd{0.0121} & \tactstd{0.0130}
& \tactstd{0.0117} & \tactstd{0.0143} & \tactstd{0.0102}
& \tactstd{0.0054} & \tactstd{0.0069} & \tactstd{0.0095}
& \tactstd{0.0073} & \tactstd{0.0082} & \tactstd{0.0101} \\[.22ex]

\cmidrule(lr){2-14}

\rowcolor{tactRowFill}
\cellcolor{tactProtocolFill} & \textbf{TACT (Full)}
& \textbf{0.9178} & \textbf{0.8794} & \textbf{0.9567}
& \textbf{0.6327} & \textbf{0.6079} & \textbf{0.7255}
& \textbf{0.7274} & \textbf{0.6867} & \textbf{0.8209}
& \textbf{0.5909} & \textbf{0.5626} & \textbf{0.6814} \\[-.4ex]

\rowcolor{tactRowFill}
\cellcolor{tactProtocolFill} & &
\tactstd{0.0006} & \tactstd{0.0007} & \tactstd{0.0005}
& \tactstd{0.0030} & \tactstd{0.0028} & \tactstd{0.0016}
& \tactstd{0.0020} & \tactstd{0.0015} & \tactstd{0.0004}
& \tactstd{0.0007} & \tactstd{0.0009} & \tactstd{0.0030} \\[.22ex]

\rowcolor{tactGainFill}
\cellcolor{tactProtocolFill}
\multirow{-25}{*}{\shortstack{Group\\LOGO}}
& {\scriptsize\emph{$\Delta$ vs.\ best (pp)}}
& +9.1 & +1.7 & +11.0
& +10.1 & +7.3 & +14.4
& +13.0 & +12.3 & +14.6
& +0.7 & +9.5 & +16.4 \\

\bottomrule
\end{tabular}%
}
\caption{Supplementary ToolBench leave-one-group-out capability ranking.
Each of the five groups is held out once and the remaining four are used
as sources. Means appear above smaller sample standard deviations over
seeds 42, 43, and 44;
best and second-best displayed means are bold and underlined. The final
row gives the absolute TACT gain over the strongest non-TACT baseline
in percentage points (pp). N@5 and H@3 denote NDCG@5 and HR@3;
Endpoint Schema denotes Endpoint Schema Match. Deterministic methods
have zero seed variance.}
\label{tab:group_held_out_results}
\end{table*}

\section{Downstream Cached Execution Details}
\label{app:cached_execution}

\subsection{Solvability Filtering and Frozen Policies}
We evaluate whether frozen capability rankings improve downstream tool
execution using the StableToolBench MirrorAPI-Cache environment.
A query is declared solvable before any ranking policy is evaluated only
when its annotated capability is present in the frozen library, at least
one permitted public endpoint can be instantiated from the available
schema, and the endpoint has a valid MirrorAPI-Cache response without
requiring authentication, a write operation, or a live external side
effect.
The final evaluation contains 120 such queries sampled from 24 ToolBench
domains, with five tasks per domain.
Table~\ref{tab:solvable_query_filter} reports the frozen
pre-evaluation filtering counts.

\begin{table*}[tp]
\centering
\small
\setlength{\tabcolsep}{5pt}
\renewcommand{\arraystretch}{1.08}
\begin{tabularx}{\textwidth}{@{}Xr@{}}
\toprule
\textbf{Pre-evaluation filter stage} & \textbf{Remaining queries} \\
\midrule
Initial candidate query pool & 5000 \\
Annotated capability is present in the frozen 3,442-capability library
& 3078 \\
At least one public \texttt{GET}/\texttt{HEAD} endpoint is available
& 1528 \\
Required arguments can be instantiated from the public schema
& 745 \\
At least one permitted endpoint has a valid MirrorAPI-Cache response
& 354 \\
No authentication, write operation, or live external side effect is
required & 147 \\
Stratified final sample: 24 domains $\times$ 5 queries
& 120 \\
\bottomrule
\end{tabularx}
\caption{
Pre-evaluation solvability filtering for cached execution.
Counts are obtained from the frozen sampling manifest before any ranking
policy is evaluated.
}
\label{tab:solvable_query_filter}
\end{table*}
For each task, we evaluate five paired capability-selection policies:
no tools, Semantic Cosine Top-5, MultiPool Top-5,
Re-Invoke Top-5, and full-model Top-5.
All ranking policies operate over the complete library of 3,442
capabilities, and their configurations are selected from source domains
only, without target relevance labels or target-domain tuning.

\subsection{Cached Execution Setting}
Recommended capabilities are converted into callable endpoints using
only the user query, public capability documentation, and public endpoint
schemas.
Only read-only \texttt{GET} and \texttt{HEAD} endpoints are allowed,
with at most 20 endpoints per case, eight endpoints per capability,
eight agent turns, and four parallel tool calls per turn.
Tool calls are executed through the local StableToolBench
MirrorAPI-Cache service, while agent and judge models are accessed
through remote LLM endpoints.
Thus, the experiment evaluates virtual StableToolBench API execution
rather than live external API services.
\subsection{Agents, Judging, and Paired Comparisons}
Four agent backbones are evaluated:
GPT-5.6 Terra, Grok-4.5, Claude Sonnet 5, and Gemini 3.1 Pro.
Each agent processes all $120 \times 5 = 600$ task--policy cases using
temperature zero and automatic function calling, producing
$120 \times 5 \times 4 = 2{,}400$ task--policy--backbone cases in total.
A blinded GPT-5.6-sol judge receives only the user request, final answer,
and tool observations; policy identities, ranking scores, recommended
capabilities, backend information, and gold annotations are hidden.
The primary metric is execution completion rate, with task-cluster bootstrap intervals, and paired exact McNemar
tests used for uncertainty analysis; results appear in
Table~\ref{tab:execution_paired}.

\begin{table*}[tp]
\centering
\small
\setlength{\tabcolsep}{4.0pt}
\renewcommand{\arraystretch}{1.08}

\begin{tabular}{lccc@{\hspace{10pt}}ccc}
\toprule
& \multicolumn{3}{c}{\textbf{Full vs. Semantic Cosine}}
& \multicolumn{3}{c}{\textbf{Full vs. Re-Invoke}} \\
\cmidrule(lr){2-4}
\cmidrule(lr){5-7}

\textbf{Agent Backbone}
& $\Delta$ Completion
& Bootstrap 95\% CI
& Exact $p$
& $\Delta$ Completion
& Bootstrap 95\% CI
& Exact $p$ \\
\midrule

GPT-5.6 Terra
& +13.33 pp
& [+5.83, +20.83]
& 0.0015
& +8.33 pp
& [+1.67, +15.83]
& 0.0414 \\

Grok-4.5
& +9.17 pp
& [+2.50, +15.83]
& 0.0127
& +8.33 pp
& [+4.17, +13.33]
& 0.0020 \\

Claude Sonnet 5
& +9.17 pp
& [+3.33, +15.83]
& 0.0074
& +8.33 pp
& [+2.50, +15.00]
& 0.0213 \\

Gemini 3.1 Pro
& +13.33 pp
& [+5.83, +20.83]
& 0.0009
& +9.17 pp
& [+2.50, +15.83]
& 0.0192 \\

\bottomrule
\end{tabular}

\caption{
Paired completion-rate comparisons of \methodname{} (Full) against
Semantic Cosine Top-5 and Re-Invoke Top-5.
Intervals are bootstrap 95\% confidence intervals, and $p$ values are
obtained from exact McNemar tests.
}
\label{tab:execution_paired}
\end{table*}

\section{Scaling and Evidence Availability Details}
\label{app:rq4_robustness}

\subsection{Candidate-Library Scaling}

Table~\ref{tab:candidate_scaling} expands the cross-domain ToolBench
candidate library from the base candidate sets to 100, 500, and all
3,442 capabilities. Pool sizes count all candidates, including positives,
rather than only newly added candidates.

For cross-dataset scaling, let $|\mathcal{C}_d|$ denote the number of
available capabilities in target dataset $d$. A nominal pool size $B$
uses $\min(B,|\mathcal{C}_d|)$ candidates; if the target library contains
fewer than 100 or 500 capabilities, the corresponding setting uses the
entire available library. Full always denotes the complete target
library. Thus, nominal sizes can coincide with Full for smaller datasets.
Seeds are averaged within targets, followed by an equal-target macro
average over ToolBench, TaskBench, MetaTool, and APIBench.

At Full scale, TACT exceeds Semantic Cosine in cross-dataset NDCG@5
by 0.0378, 0.0478, 0.0572, and 0.0469 on Natural, Semantic, Context,
and Mixed, respectively. The equal-view averages are 0.4475 for TACT
and 0.4001 for Semantic Cosine, a difference of 0.0474. These results
describe recovery of annotated positives from expanded libraries and
do not imply a smaller absolute Base-to-Full decline.

\subsection{Fixed-Checkpoint Source-Evidence Diagnostic}

The source-evidence intervention uses the Full checkpoints from the
primary ranking experiment under the corresponding transfer protocol.
The outer splits, selected parameters, calibration, and candidate pools
are fixed. Checkpoints and calibration are shared across Natural,
Semantic, Context, and Mixed. Only the fraction of retained source
associations changes; the model is not retrained or reselected for an
evidence level.

At each level, Full incorporates all available source evidence, whereas \emph{Fixed-Checkpoint Functional Fallback} excludes the contextual contribution while keeping the remaining score components unchanged. When no source evidence is available, the two settings become identical. Cross-domain results are macro-averaged over the same eligible target domains as in the primary ranking experiment, while cross-dataset results are macro-averaged across the four target datasets.

This intervention measures the contribution of contextual correction
within the fitted Full model. The RQ2 w/o Context ablation instead
trains and selects a separate model without that branch. The former is
a fixed-checkpoint diagnostic; the latter is an independently trained
branch ablation.

\subsection{Closed-World Evaluation Boundary}

Expanded pools retain the frozen relevance boundary: annotated
positives are relevant and additional unjudged capabilities are treated
as non-relevant. Scaling measures recovery of known positives, not
the incompatibility of every unannotated capability. Target labels
define evaluation sets and final metrics; they are not used for
training, selection, expansion ranking, or evidence interventions.

\clearpage
\endgroup
\end{document}